\documentclass[10pt,twocolumn,letterpaper]{article}

\usepackage{main/iccv}
\usepackage{times}
\usepackage{epsfig}
\usepackage{graphicx}
\usepackage{amsmath}
\usepackage{amssymb}
\usepackage{stix}

\usepackage{booktabs}
\usepackage{multirow}
\usepackage{multicol}

\usepackage{pifont}

\usepackage{enumitem}
\usepackage{microtype}
\usepackage{subcaption}
\usepackage{flushend}

\usepackage{balance}

\usepackage{contour}
\contourlength{0.33pt}

\usepackage{pgf,tikz}
\usetikzlibrary{shapes.geometric}
\usepackage{pgfplots}
\usepackage{filecontents}
\pgfplotsset{compat=1.7}
\newlength\figureheight
\newlength\figurewidth

\usepackage{soul}

\usepackage[permil]{overpic}

\usepackage{todonotes}

\usepackage{wrapfig}

\newcommand{\ourmethod}{FCGF6D\xspace}

\definecolor{ape}{RGB}{143,24,24}
\definecolor{can}{RGB}{237,154,59}
\definecolor{cat}{RGB}{145,70,163}
\definecolor{driller}{RGB}{57,67,53}
\definecolor{duck}{RGB}{226,215,62}
\definecolor{eggbox}{RGB}{111,161,18}
\definecolor{glue}{RGB}{150,141,137}
\definecolor{holepuncher}{RGB}{46,54,110}

\definecolor{master chef can}{RGB}{16, 16, 230}
\definecolor{sugar box}{RGB}{93, 16, 235}
\definecolor{tomato soup can}{RGB}{13, 212, 142}
\definecolor{tuna fish can}{RGB}{111, 161, 18}
\definecolor{potted meat can}{RGB}{15, 214, 44}
\definecolor{bleach cleanser}{RGB}{16, 166, 235}
\definecolor{bowl}{RGB}{232, 225, 23}
\definecolor{mug}{RGB}{237, 7, 26}
\definecolor{wood block}{RGB}{43, 16, 4}
\definecolor{scissors}{RGB}{117, 87, 26}
\definecolor{large clamp}{RGB}{218, 9, 237}
\definecolor{extra large clamp}{RGB}{6, 54, 13}

\usepackage[pagebackref=true,breaklinks=true,letterpaper=true,colorlinks,bookmarks=false]{hyperref}

\usepackage[capitalize]{cleveref}
\crefname{section}{Sec.}{Secs.}
\Crefname{section}{Section}{Sections}
\Crefname{table}{Table}{Tables}
\crefname{table}{Tab.}{Tabs.}

\def\iccvPaperID{3}

\ificcvfinal\fi
\iccvfinalcopy

\begin{document}

\title{Revisiting Fully Convolutional Geometric Features for Object 6D Pose Estimation}

\author{
\begin{minipage}{0.32\textwidth}
    \vspace*{-7mm}
    \centering
    Jaime Corsetti
\end{minipage}
\begin{minipage}{0.32\textwidth}
    \vspace*{-7mm}
    \centering
    Davide Boscaini
\end{minipage}
\begin{minipage}{0.32\textwidth}
    \vspace*{-7mm}
    \centering
    Fabio Poiesi
\end{minipage} \\
\begin{minipage}{0.32\textwidth}
    \vspace*{-3mm}
    \centering
    {\tt\small jaime.corsetti98@gmail.com}
\end{minipage}
\begin{minipage}{0.32\textwidth}
    \vspace*{-3mm}
    \centering
    {\tt\small dboscaini@fbk.eu}
\end{minipage}
\begin{minipage}{0.32\textwidth}
    \vspace*{-3mm}
    \centering
    {\tt\small poiesi@fbk.eu}
\end{minipage} \\
\begin{minipage}{1.0\textwidth}
    \vspace*{1mm}
    \centering
    Fondazione Bruno Kessler, Italy
\end{minipage}
}

% Jaime Corsetti \hspace{1.7cm} Davide Boscaini \hspace{1.7cm} Fabio Poiesi\\
% Technologies of Vision Lab, Fondazione Bruno Kessler, Trento, Italy \\
% {\tt\small \hspace{-1.3cm} jaime.corsetti98@gmail.com \hspace{0.5cm} dboscaini@fbk.eu \hspace{1.5cm} poiesi@fbk.eu}

\maketitle

\begin{abstract}
    Recent works on 6D object pose estimation focus on learning keypoint correspondences between images and object models, and then determine the object pose through RANSAC-based algorithms or by directly regressing the pose with end-to-end optimisations.
    We argue that learning point-level discriminative features is overlooked in the literature.
    To this end, we revisit Fully Convolutional Geometric Features (FCGF) and tailor it for object 6D pose estimation to achieve state-of-the-art performance.
    FCGF employs sparse convolutions and learns point-level features using a fully-convolutional network by optimising a hardest contrastive loss.
    We can outperform recent competitors on popular benchmarks by adopting key modifications to the loss and to the input data representations, by carefully tuning the training strategies, and by employing data augmentations suitable for the underlying problem.
    We carry out a thorough ablation to study the contribution of each modification.
    The code is available at \url{https://github.com/jcorsetti/FCGF6D}.
\end{abstract}

%%%%%%%%%%%%%%%%%%%%%%%%%%%%%%%%%%%%%%%%%%%%%%%%%%%%%%%%%%%%%%%%%%
%%%%%%%%%%%%%%%%%%%%%%%%%%%%%%%%%%%%%%%%%%%%%%%%%%%%%%%%%%%%%%%%%%
%%%%%%%%%%%%%%%%%%%%%%%%%%%%%%%%%%%%%%%%%%%%%%%%%%%%%%%%%%%%%%%%%%
\section{Introduction}\label{sec:intro}

Object 6D pose estimation is the problem of finding the Euclidean transformation (i.e.~pose) of an object in a scene with respect to the camera frame~\cite{hodavn2016evaluation}.
This problem is important for autonomous driving~\cite{autodriving1}, augmented reality~\cite{augreality}, space docking~\cite{space}, robot grasping~\cite{grasping1}, and active 3D classification \cite{Wang2019}.
The main challenges are handling occlusions, structural similarities between objects, and non-informative textures.
Different benchmarks have been designed to study these challenges, such as LineMod-Occluded (LMO)~\cite{lmo}, YCB-Video (YCBV)~\cite{ycbv}, and T-LESS~\cite{tless}. 
LMO includes poorly-textured objects in scenarios with several occlusions.
In YCBV, well-textured objects appear in scenarios with fewer occlusions but more pose variations.
T-LESS includes poorly-textured and geometrically-similar objects in industrial scenarios with occasional occlusions.

% ================================================================
\begin{figure}[t]
    \centering
    \input{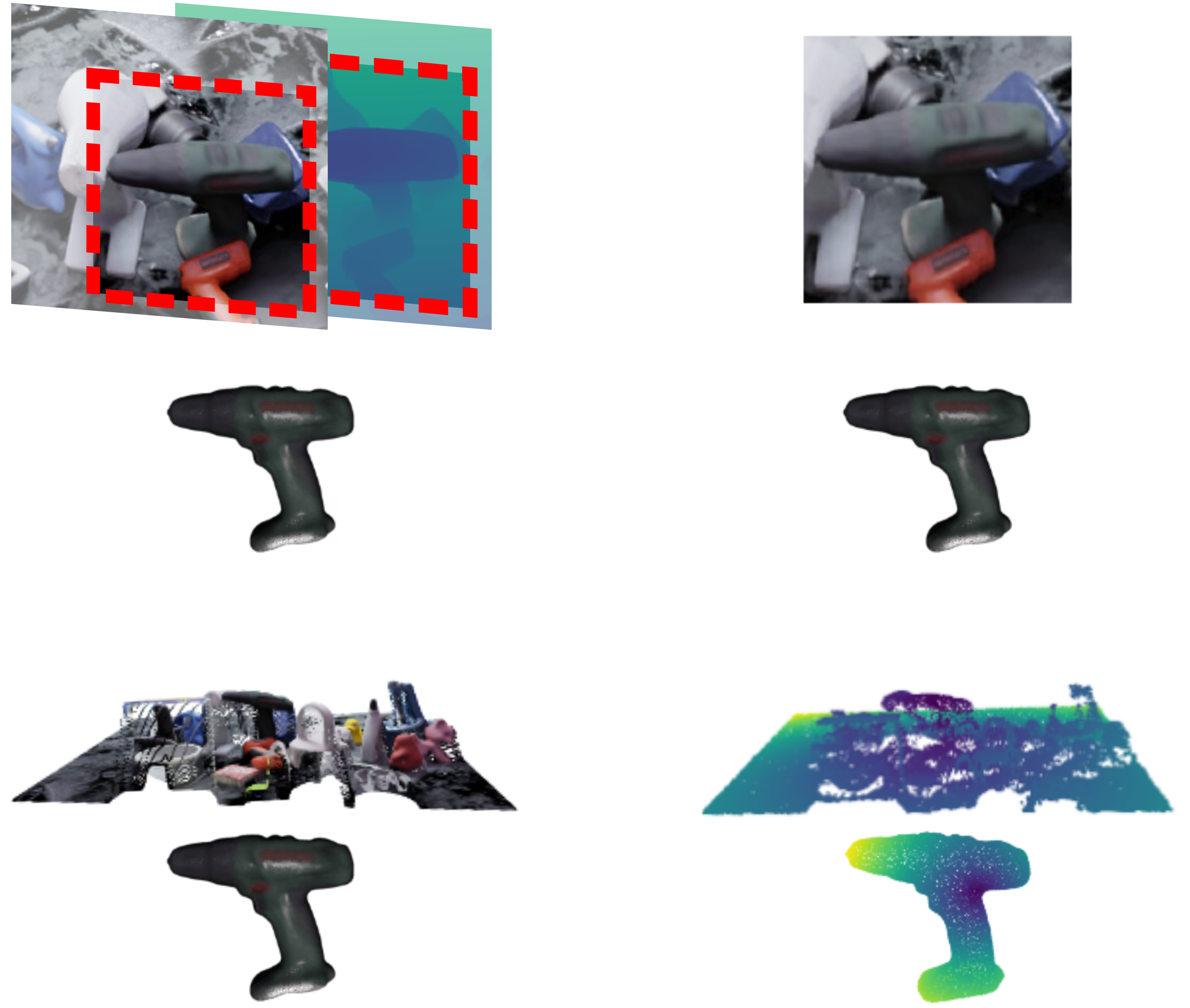}
    \vspace*{-7mm}
    \caption{
    Top: Typically, two-stage 6D pose estimation methods process the input (RGBD image, 3D object) with different deep neural networks (2D, 3D) to learn keypoint correspondences~\cite{geometricaware6d}, or directly predict the keypoint projections on the image~\cite{pvn3d,ffb6d}.
    They also rely on detectors to crop the input image, and estimate the final pose with RANSAC-based PnP~\cite{ransac}.
    Bottom: Our method processes the whole scene and the object point clouds with 3D deep neural networks, optimises the output point-wise (dense) features by using ground-truth correspondences, and estimates the final pose with a point cloud registration algorithm.
    }
    \label{fig:teaser}
\end{figure}
% ================================================================

Object 6D pose estimation approaches based on deep learning can be classified as \textit{one-stage}~\cite{singlestage,e2ek,deepim} or \textit{two-stage}~\cite{segdriven,pvn3d,ffb6d,geometricaware6d}.
One-stage approaches can directly regress the object pose~\cite{singlestage,e2ek,deepim}.
Two-stage approaches can predict 3D keypoints~\cite{pvn3d,ffb6d} or point-level correspondences between the scene and the object~\cite{geometricaware6d}.
Correspondences can be computed through point-level features~\cite{geometricaware6d}.
One-stage approaches are typically more efficient than their two-stage counterpart, as they require only one inference pass. However, rotation regression is a difficult optimisation task because the rotation space is non-Euclidean and non-linear, and the definition of correct orientation is ambiguous in case of symmetric objects~\cite{saxena2009learning}.
On the other hand, correspondence-based approaches have to be coupled with registration techniques, such as RANSAC, PnP, or least square estimation~\cite{geometricaware6d}.

We argue that the problem of learning discriminative point-level features is overlooked in the related literature.
Moreover, we believe that working at intermediate levels of representation learning, rather than regressing the pose directly, facilitates interpretability and enables us to effectively debug algorithms.
Literature on representation learning for point cloud registration has made great advances~\cite{FCGF,GEDI}, and none of the object 6D pose estimation methods have deeply investigated the application of these techniques to the underlying problem (Fig.~\ref{fig:teaser}).
In a landscape dominated by complex networks, our work stands as the first to comprehensively explore and quantify the benefits of this formulation with a simple yet effective solution. Our research addresses fundamental and previously unanswered questions:\\
\textit{i) How to learn features of heterogeneous point clouds (objects and scenes) that align in the same representation space and exhibit cross-domain generalisation (synthetic to real)?\\
ii) What training strategies are optimal for this approach?\\
iii) What degree of improvement can these strategies bring?}
To answer these questions, we revisit Fully Convolutional Geometric Features (FCGF)~\cite{FCGF} and show that its potential to achieve state-of-the-art results lies in an attentive design of data augmentations, loss negative mining, network architecture, and optimisation strategies.
FCGF is designed to learn point-level features by using a fully-convolutional network optimised through a hardest contrastive loss.
Compared to the original FCGF setting, our setting is asymmetric, i.e.~the two input point clouds have different sizes and resolutions.
Therefore, we modify the hardest contrastive loss to take into account the size of each point cloud for the mining of the hardest negatives.
We use separate architectures to learn specific features for the two (heterogeneous) input data (object and scene), but unlike several state-of-the-art methods we train only a single model for all the objects of each dataset.
We use specific augmentations to tackle occlusions, which are the main challenge in real-world scenarios and in the considered datasets.
We name our approach \ourmethod.
\ourmethod outperforms state-of-the-art methods (+3.5 ADD(S)-0.1d on LMO, +0.8 ADD-S AUC on YCBV), even when comparing with methods that train one model for each object.
Our ablation study suggests that most of the performance gain is obtained thanks to our changes to the loss, the addition of the RGB information and our changes to the optimizer.
In summary, our contributions are:
\setlist{nolistsep}
\begin{itemize}
    \item We tailor FCGF for object 6D pose estimation in order to 
    i) process entire scenes rather than cropped regions as competitors,
    ii) learn a single model for all objects instead of a model for each object,
    iii) process both photometric and geometric information with a single unified deep network model.
    \item A modified version of the hardest contrastive loss that is applied to heterogeneous point clouds and that considers a geometric constraint when mining the hardest negative.
    \item We study data augmentations that enable FCGF to improve generalisation between synthetic and real data.
\end{itemize}

%%%%%%%%%%%%%%%%%%%%%%%%%%%%%%%%%%%%%%%%%%%%%%%%%%%%%%%%%%%%%%%
%%%%%%%%%%%%%%%%%%%%%%%%%%%%%%%%%%%%%%%%%%%%%%%%%%%%%%%%%%%%%%%
%%%%%%%%%%%%%%%%%%%%%%%%%%%%%%%%%%%%%%%%%%%%%%%%%%%%%%%%%%%%%%%
\section{Related work}\label{sec:related}

6D pose estimation approaches can be designed to use different input data.
RGB methods~\cite{segdriven,singlestage,sopose,zebrapose} rely on photometric information only, while RGBD methods~\cite{pvn3d,ffb6d,surfemb,e2ek,geometricaware6d} also use range information in addition to RGB.

%%%%%%%%%%%%%%%%%%%%%%%%%%%%%%%%%%%%%%%%%%%%%%%%%%%%%%%%%%%%%%%
\noindent \textbf{RGB-based 6D pose estimation.}
SO-Pose~\cite{sopose} proposes an end-to-end method that explicitly models self-occlusion maps (i.e., portions of the object that are hidden by camera orientation). 
It computes 2D-3D correspondences for each visible point of the object, and feeds them with self-occlusion maps to a pose regression module.
ZebraPose~\cite{zebrapose} proposes a strategy to learn surface descriptors on the image, by training a neural network to predict pixel features which correspond to predefined descriptors on the object model. 
At inference time, it finds correspondences by similarity, and solves the PnP problem with RANSAC.
The authors show that the vertex encoding process is crucial for performance improvement. 

%%%%%%%%%%%%%%%%%%%%%%%%%%%%%%%%%%%%%%%%%%%%%%%%%%%%%%%%%%%%%%%
\noindent \textbf{RGBD-based 6D pose estimation.}
PVN3D~\cite{pvn3d} extends PVNet~\cite{pvnet} by incorporating 3D point cloud information. 
The core of this approach is a keypoint voting mechanism, in which for each pixel the offset to a reference keypoint is regressed.
A semantic segmentation module is also used to identify the points belonging to each object in the scene.
PVN3D is a two-stage method, as it passes the final correspondences to a RANSAC-based~\cite{ransac} algorithm for 6D pose estimation.
FFB6D~\cite{ffb6d} adopts an analogous method to PVN3D~\cite{pvn3d}, but introduces a novel convolutional architecture with Fusion Modules.
These modules enable the model to combine photometric (RGB) and geometrical (D) features for learning a better point cloud representation.
E2EK~\cite{e2ek} proposes an end-to-end trainable method by extending FFB6D~\cite{ffb6d}. 
It clusters and filters the features computed by FFB6D based on confidence, and then processes them by an MLP-like network that regresses the pose. 
Wu et al.~\cite{geometricaware6d} addresses the problem of objects that are symmetric to rotation with a two-stage method. 
They extend FFB6D~\cite{ffb6d} by introducing a novel triplet loss based on geometric consistency. 
Symmetry is leveraged by considering symmetric points as positives, thus forcing them to have similar features.
Feng et al.~\cite{feng2020fully} proposes a method to solve a related problem. In this work, FCGF is applied to align different point clouds of objects belonging to the same category. However, the authors do not introduce task-specific modifications to FCGF, and unlike our case of application, the target object is assumed to be already segmented from the scene.

Unlike methods that employ sophisticated combinations of deep network architectures to process RGB and depth modalities \cite{ffb6d,geometricaware6d}, our approach
uses deep networks based on sparse convolutions to process coloured point clouds with a single framework.
Sparse convolutions are designed to process point clouds efficiently \cite{minkowski}.
We also split the pose estimation problem into two subproblems, i.e.~feature learning and point cloud registration.
This allows us to evaluate the quality of the learned features by using metrics such as Feature Matching Recall~\cite{deng2018ppfnet}, which fosters interpretability of our model. 
Unlike Wu et al.~\cite{geometricaware6d}, we do not rely on a detector to crop the region with the candidate object before processing the point cloud with our network.
Our experiments show that we can outperform the nearest competitors E2EK~\cite{e2ek} and Wu et al.~\cite{geometricaware6d} by 5.7 and 1.9 ADD(S)-0.1d on the LMO dataset, respectively, without using a detector.

% =======================================================================
\begin{figure*}[t]
    \centering
    \includegraphics[width=0.9\textwidth]{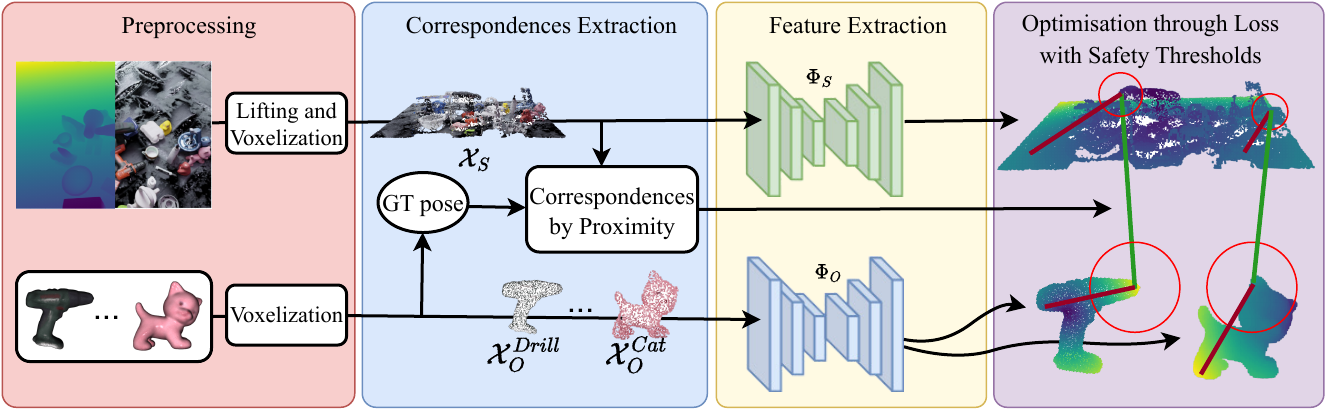}
    \vspace{-2mm}
    \caption{
    \ourmethod training pipeline consists of four logical parts.
    Given a scene $S$ and an object $O$ we take as input the pair $(\mathcal{I}_S, \mathcal{M}_O)$.
    In the first part, we compute 3D point cloud representations $(\mathcal{X}_S, \mathcal{X}_O)$ of $(\mathcal{I}_S, \mathcal{M}_O)$, where $\mathcal{X}_S$ is obtained by lifting $\mathcal{I}_S$ using the intrinsic parameters of the camera that acquired it, and then quantise them.
    In the second part, we mine positives by computing the correspondences between $\mathcal{X}_S$ and $ \widetilde{\mathcal{X}}_O = \textbf{R}_O \mathcal{X}_O + \textbf{t}_O$, where $\textbf{R}_O, \textbf{t}_O$ is the ground-truth 6D pose of $O$.
    In the third part, we perform point-wise feature extraction with two independent UNets $\Phi_S, \Phi_O$.
    In the fourth part, the hardest contrastive loss with safety thresholds is applied to guide the feature learning process.
    }
    \label{fig:pipeline}
\end{figure*}
% =======================================================================
%%%%%%%%%%%%%%%%%%%%%%%%%%%%%%%%%%%%%%%%%%%%%%%%%%%%%%%%%%%%%%%%%%%%%%%%%
%%%%%%%%%%%%%%%%%%%%%%%%%%%%%%%%%%%%%%%%%%%%%%%%%%%%%%%%%%%%%%%%%%%%%%%%%
%%%%%%%%%%%%%%%%%%%%%%%%%%%%%%%%%%%%%%%%%%%%%%%%%%%%%%%%%%%%%%%%%%%%%%%%%
\section{Preliminary: A review of FCGF}

%%%%%%%%%%%%%%%%%%%%%%%%%%%%%%%%%%%%%%%%%%%%%%%%%%%%%%%%%%%%%%%%%%%%%%%%%
\noindent \textbf{Input data representation.}
FCGF takes as input a quantised version of the original point cloud $\mathcal{X} \in \mathbb{R}^{V \times 3}$.
The quantisation procedure splits the volume occupied by $\mathcal{X}$ into a grid of voxels of size $Q$ and assigns a single representative vertex $\textbf{x}_i \in \mathbb{R}^3$ to each voxel $i$.
This reduction is typically computed with random sampling or by average pooling (barycenter)~\cite{minkowski}.
The resulting sparse representation is obtained by discarding voxels corresponding to a portion of the empty space and is significantly more efficient in terms of memory utilisation.

%%%%%%%%%%%%%%%%%%%%%%%%%%%%%%%%%%%%%%%%%%%%%%%%%%%%%%%%%%%%%%%%%%%%%%%%%
\noindent \textbf{Feature extractor.}
The fully-convolutional feature extractor $\boldsymbol{\Phi}_\Theta$ is a parametric function with learnable parameters $\Theta$ designed as a UNet~\cite{UNet}.
Given $\textbf{x}_i$, $\boldsymbol{\Phi}_\Theta$ produces a $F$-dimensional feature vector defined as $\boldsymbol{\Phi}_\Theta( \textbf{x}_i ) = \textbf{f}_i \in \mathbb{R}^F$.
FCGF processes pairs of point clouds using a Siamese approach, i.e.~feature extractors with shared weights.
% The weight sharing implicitly enforces symmetry in the metric learned by the model.
FCGF is implemented in PyTorch using Minkowski engine~\cite{minkowski}.

%%%%%%%%%%%%%%%%%%%%%%%%%%%%%%%%%%%%%%%%%%%%%%%%%%%%%%%%%%%%%%%%%%%%%%%%%
\noindent \textbf{Hardest contrastive loss.}
The hardest contrastive (HC) loss is defined as $\ell_\text{HC} = \lambda_P \ell_P + \lambda_N \ell_N$, where $\ell_P$ promotes similarity between features of positive samples, $\ell_N$ promotes dissimilarity between features of negative samples, and $\lambda_P, \lambda_N$ are hyperparameters. Given a pair of 3D scenes $( \mathcal{X}_1, \mathcal{X}_2 )$ as input, the set of positive pairs is defined as $\mathcal{P} = \{ (i, j): \textbf{x}_i \in \mathcal{X}_1, \textbf{x}_j \in \mathcal{X}_2, \phi(\textbf{x}_i) = \textbf{x}_j \}$, where $\phi \colon \mathcal{X}_1 \to \mathcal{X}_2$ is a correspondence mapping between $\mathcal{X}_1$ and $\mathcal{X}_2$ voxels.
% Positive term
$\ell_P$ is defined as
%+++++++++++++++++++++++++++++++++++++++
%\vspace{-1.5mm}
\begin{equation}\label{eq:hcpos}
\ell_P = \sum_{(i, j) \in \mathcal{P}} \frac{1}{\lvert \mathcal{P} \rvert} \left( \lVert \textbf{f}_i - \textbf{f}_j \rVert - \mu_P \right)^2_+,
%\vspace{-3mm}
\end{equation}
%+++++++++++++++++++++++++++++++++++++++
where $\lvert \mathcal{P} \rvert$ is the cardinality of $\mathcal{P}$, $\mu_P$ is a positive margin to overcome overfitting~\cite{lin2015deephash}, and $( \cdot )_+ = \max(0, \cdot)$.
% Negative mining
For each pair $(i, j) \in \mathcal{P}$, two sets of candidate negatives are defined as $\mathcal{N}_i = \{ k \text{ s.t. } \textbf{x}_k \in \mathcal{X}_1, k \ne i \}$, $\mathcal{N}_j = \{ k \text{ s.t. } \textbf{x}_k \in \mathcal{X}_2, k \ne j \}$.
Computing $\mathcal{N}_i, \mathcal{N}_j$ scales quadratically with the minibatch size, therefore random subsets of $\mathcal{N}_i$ and $\mathcal{N}_j$ with fixed cardinalities are instead used in practice.
% Negative term
$\ell_N$ is defined as
\begin{equation}\label{eq:hcneg}
\begin{aligned}
\vspace{-2mm}
\ell_N = \sum_{(i, j) \in \mathcal{P}} & \frac{1}{2 \lvert \mathcal{P}_i \rvert} \left( \mu_N - \min_{k \in \mathcal{N}_i} \lVert \textbf{f}_i - \textbf{f}_k \rVert \right)^2_+ \\
& \hspace{-4.5mm}+ \frac{1}{2 \lvert \mathcal{P}_j \rvert} \left( \mu_N - \min_{k \in \mathcal{N}_j} \lVert \textbf{f}_j - \textbf{f}_k \rVert \right)^2_+,
\end{aligned}
\end{equation}
where $\lvert \mathcal{P}_i \rvert, \lvert \mathcal{P}_j \rvert$ are the numbers of valid negatives mined from the first and second term, respectively.
Unlike metric learning losses that randomly mine a certain number of negatives from $\mathcal{N}_i, \mathcal{N}_j$~\cite{ContrastiveLoss, TripletLoss}, the HC loss mines the most similar features within a batch, i.e.~the hardest negatives.

%%%%%%%%%%%%%%%%%%%%%%%%%%%%%%%%%%%%%%%%%%%%%%%%%%%%%%%%%%%%%%%%%%%%%%%%%%%%%
%%%%%%%%%%%%%%%%%%%%%%%%%%%%%%%%%%%%%%%%%%%%%%%%%%%%%%%%%%%%%%%%%%%%%%%%%%%%%
%%%%%%%%%%%%%%%%%%%%%%%%%%%%%%%%%%%%%%%%%%%%%%%%%%%%%%%%%%%%%%%%%%%%%%%%%%%%%
\section{Tailoring FCGF for 6D pose estimation}
In this section, we describe how we modified FCGF. We focus on manipulating heterogeneous representations of input data, improving the HC loss, and modernising the training strategy. 
Fig.~\ref{fig:pipeline} shows the block diagram of \ourmethod.

%%%%%%%%%%%%%%%%%%%%%%%%%%%%%%%%%%%%%%%%%%%%%%%%%%%%%%%%%%%%%%%%%%%%%%%%%%%%%
\subsection{Input data}

\noindent \textbf{Heterogeneous representations.}
FCGF was designed for scene registration, where its input data is 3D scan pairs of the same scene captured from different viewpoints.
Therefore, their input data belongs to the same distribution, i.e.~real-world data captured with the same LiDAR sensor.
This is why authors in \cite{FCGF} use a Siamese approach.
Unlike FCGF, our input data is heterogeneous, therefore we process it with two independent deep networks. Formally, given an object $O$ and a scene $S$, the input of our pipeline is the pair $(\mathcal{M}_O, \mathcal{I}_S)$, where $\mathcal{M}_O$ is a textured 3D model of $O$ and $\mathcal{I}_S$ is an RGBD capture of $S$ from a viewpoint.
We transform $(\mathcal{M}_O, \mathcal{I}_S)$ into a pair of point clouds.
For $O$, we produce a point cloud $\mathcal{X}_O \in \mathbb{R}^{V_O\times6}$ by sampling $V_O$ vertices on the triangular faces of $\mathcal{M}_O$ and extracting the corresponding RGB colours from its texture.
For $S$, we use the intrinsic parameters of the RGBD sensor to map $\mathcal{I}_S$ into a coloured point cloud and sample $V_S$ points from it.
Let $\mathcal{X}_S \in \mathbb{R}^{V_S\times6}$ be the point cloud of $S$.
We quantise $\mathcal{X}_O$ and $\mathcal{X}_S$ by a factor $Q$ and process the pair with two networks implemented with Minkowski engine~\cite{minkowski}.
$V_O$, $V_S$, and $Q$ are hyperparameters.

%%%%%%%%%%%%%%%%%%%%%%%%%%%%%%%%%%%%%%%%%%%%%%%%%%%%%%%%%%%%%%%%%%%%%%%%%%%%%
\noindent \textbf{Processing geometric and photometric data.}
Minkowski engine~\cite{minkowski} is designed to process optional input features in addition to the 3D coordinate of each point.
However, authors in \cite{FCGF} show that, in the context of scene registration, adding the photometric information associated to each point leads to overfitting.
We found instead that this addition significantly improves the performance. 
Colour information helps in i) discriminating objects of different categories but with similar geometric shape (e.g. pudding box and gelatin box in YCBV~\cite{ycbv}), and ii) selecting the correct pose of symmetric objects among the set of geometrically-equivalent ones (i.e. the 6D pose of a box or a can cannot be uniquely defined unless we consider their texture patterns).

%%%%%%%%%%%%%%%%%%%%%%%%%%%%%%%%%%%%%%%%%%%%%%%%%%%%%%%%%%%%%%%%%%%%%%%%%%%%%
%%%%%%%%%%%%%%%%%%%%%%%%%%%%%%%%%%%%%%%%%%%%%%%%%%%%%%%%%%%%%%%%%%%%%%%%%%%%%
\subsection{Loss function}

%%%%%%%%%%%%%%%%%%%%%%%%%%%%%%%%%%%%%%%%%%%%%%%%%%%%%%%%%%%%%%%%%%%%%%%%%%%%%
\noindent \textbf{Positive mining.}
We define $\mathcal{P}$ (Eq.~\ref{eq:hcpos}) as the set of valid correspondences between $\mathcal{X}_O$ and $\mathcal{X}_S$.
Let $(\mathbf{R}_O, \mathbf{t}_O)$ be $O$ ground-truth 6D pose in $S$ and $\widetilde{\mathcal{X}}_O = \mathbf{R}_O \mathcal{X}_O + \mathbf{t}_O$ be the rigidly transformed version of $\mathcal{X}_O$ into the reference frame of $\mathcal{X}_S$.
We compute all the correspondences by searching for each point of $\widetilde{\mathcal{X}}_O$ its nearest neighbouring point in $\mathcal{X}_S$.
Due to occlusions with other objects and/or self-occlusions, some of the correspondences may be spurious, e.g.~associating points of different surfaces.
Therefore, we consider a correspondence valid if the distance between $\tilde{\textbf{x}}_i \in \widetilde{\mathcal{X}}_O$ and $\textbf{x}_j \in \mathcal{X}_S$ is less than a threshold $\tau_P$ and if the other points on the scene are farther away, i.e. $(i, j) \in \mathcal{P} \Leftrightarrow \lVert \tilde{\textbf{x}}_i - \textbf{x}_j \rVert < \tau_P \text{ and } \lVert \tilde{\textbf{x}}_i - \textbf{x}_j \rVert < \lVert \tilde{\textbf{x}}_i - \textbf{x}_k \rVert $ for every $k = 1, \dots, V_S$.
%%%%%%%%%%%%%%%%%%%%%%%%%%%%%%%%%%%%%%%%%%%%%%%%%%%%%%%%%%%%%%%%%%%%%%%%%%%%%
\noindent \textbf{Negative mining.}
We experienced that mining the hardest negatives from the negative sets $\mathcal{N}_i$, $\mathcal{N}_j$ (Eq.~\ref{eq:hcneg}) can lead to loss instability and collapsing. % experimentally 
This occurs because the hardest negative in $\mathcal{N}_i = \{ k : \textbf{x}_k \in \mathcal{X}_O, k \ne i \}$, i.e. the sample with the closest feature to $\mathbf{f}_i \in \mathbb{R}^F$, is likely to be a point spatially close to $\textbf{x}_i \in \mathcal{X}_O$, because their local geometric structure is nearly the same.
Hence, Eq.~\ref{eq:hcneg} tries to enforce features corresponding to the same local geometric structure to be distant from each other.
This problem can be mitigated by replacing $\mathcal{N}_i, \mathcal{N}_j$ in Eq.~\ref{eq:hcneg} with $\widetilde{\mathcal{N}}_i = \{ k : \textbf{x}_k \in \mathcal{X}_O, \lVert \textbf{x}_k - \textbf{x}_i \rVert > \tau_{NO} \}$ and $\widetilde{\mathcal{N}}_j = \{ k : \textbf{x}_k \in \mathcal{X}_S, \lVert \textbf{x}_k - \textbf{x}_j \rVert > \tau_{NO} \}$, where $\tau_{NO}$ is a safety threshold, i.e.~the radius of spheres on object and on scene where mining is forbidden.

\begin{figure*}[t]
\input{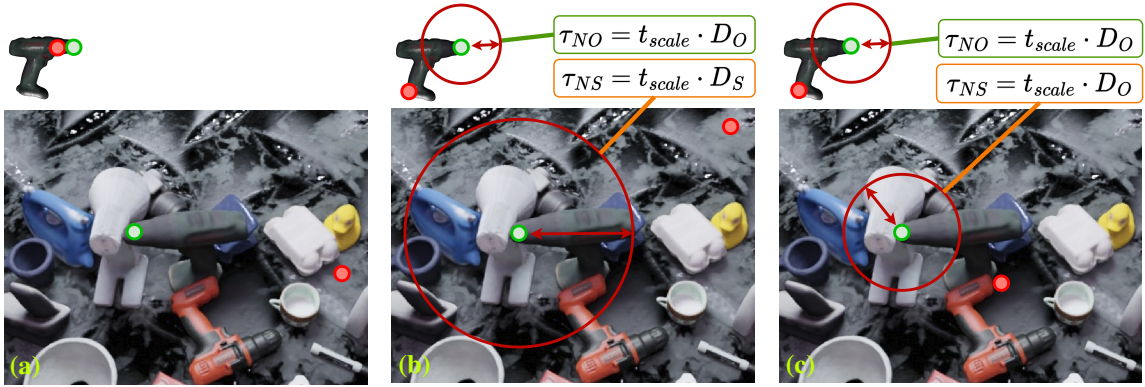}
\vspace*{-1mm}
\caption{
    Examples of different mining strategies. 
    (a) Hardest contrastive loss as proposed in FCGF: no constraints are enforced on the location of the hardest negative (red point) with respect to the correspondent point (green point).
    (b) A vanilla choice of the safety thresholds: the radii $\tau_{NS}$, $\tau_{NO}$ are proportional to the diameters $D_{S}$, $D_{O}$ of the respective point clouds.
    (c) Our choice: the value of the thresholds is proportional to the diameter of the object, i.e.~$\tau_{NS} = \tau_{NO}$.
}
\label{fig:loss_detail}
\end{figure*}

The choice of $\tau_{NO}$ is key because it determines which points on the point clouds can be used for negative mining.
We found beneficial to choose $\tau_{NO}$ as a function of the dimension of the input object.
Given $\mathcal{X}_O$, we define its diameter as $D_O$, and set $\tau_{NO}=t_{\text{scale}} D_O$. 
In Fig.~\ref{fig:loss_detail}, we illustrate the safety thresholds.
In this way, we can maintain a good quantity of negatives while avoiding the mining of spurious hardest negatives. 
Using different thresholds for the object and the scene points clouds underperformed our final choice.
Therefore, our loss is defined as
%++++++++++++++++++++++++++++++++++
\begin{equation*}
    \ell_\text{HC} = \lambda_P \ell_P + \lambda_{NO} \ell_{NO} + \lambda_{NS} \ell_{NS},
\end{equation*}
%+++++++++++++++++++++++++++++++++++
where $\lambda_{P}$, $\lambda_{NO}$ and $\lambda_{NS}$ are weight factors. $\tau_P$, $t_{\text{scale}}$, $\lambda_{P}$, $\lambda_{NO}$, and $\lambda_{NS}$ are hyperparameters.

%===================================
%===================================

%%%%%%%%%%%%%%%%%%%%%%%%%%%%%%%%%%%%%%%%%%%%%%%%%%%%%%%%%%%%%%%%%%%%%%%%%%%%%
%%%%%%%%%%%%%%%%%%%%%%%%%%%%%%%%%%%%%%%%%%%%%%%%%%%%%%%%%%%%%%%%%%%%%%%%%%%%%
\subsection{Training strategy}

%%%%%%%%%%%%%%%%%%%%%%%%%%%%%%%%%%%%%%%%%%%%%%%%%%%%%%%%%%%%%%%%%%%%%%%%%%%%%
\noindent \textbf{Data augmentation.}
FCGF combines scaling and rotation augmentations to enhance feature robustness against variations in camera pose~\cite{FCGF}.
These are effective in the context of point cloud registration, but in our specific scenario, the point cloud of the objects always belongs to a known set.
Avoiding these augmentations helps the deep network in learning specialised features for each object. Our data augmentations consist of the following:

(i) Point re-sampling of $O$ and $S$, i.e.~unlike FCGF, we randomly downsample point clouds at each epoch to mitigate overfitting. This allows the model to be more robust to depth acquisition noise; 
(ii) Colour jittering on $O$, i.e.~we randomly perturb brightness, contrast, saturation, and hue of $O$;
(iii) Random erasing on $S$, i.e.~unlike FCGF, we simulate occlusions at training time. For each point of $\widetilde{\mathcal{X}}_O$ we compute its nearest neighbour in $\mathcal{X}_S$ and randomly select a point on $\mathcal{X}_S$ within such correspondence set. We then erase all the points that fall within a distance threshold $\rho$ from it. This allows the model to be more robust to occlusions in the input scene.

%%%%%%%%%%%%%%%%%%%%%%%%%%%%%%%%%%%%%%%%%%%%%%%%%%%%%%%%%%%%%%%%%%%%%%%%%%%%%
\noindent \textbf{Optimisation techniques.} 
FCGF uses an SGD optimiser with an initial learning rate $\text{lr}_{\text{init}}=10^{-1}$ decreased during training with an exponential scheduler with $\gamma = 0.99$.
In our setting, these hyperaparameters do not lead to convergence.
Instead, we set $\text{lr}_{\text{init}}=10^{-3}$.
We experiment with Adam~\cite{adam} and AdamW~\cite{adamw}, and notice improvements in both cases.
We also switch to a Cosine Annealing scheduler~\cite{cosine} that lowers the learning rate from $10^{-3}$ to $10^{-4}$ across the epochs.

%%%%%%%%%%%%%%%%%%%%%%%%%%%%%%%%%%%%%%%%%%%%%%%%%%%%%%%%%%%%%%%%%%%%
%%%%%%%%%%%%%%%%%%%%%%%%%%%%%%%%%%%%%%%%%%%%%%%%%%%%%%%%%%%%%%%%%%%%
%%%%%%%%%%%%%%%%%%%%%%%%%%%%%%%%%%%%%%%%%%%%%%%%%%%%%%%%%%%%%%%%%%%%
\section{Experiments}\label{sec:results}

%%%%%%%%%%%%%%%%%%%%%%%%%%%%%%%%%%%%%%%%%%%%%%%%%%%%%%%%%%%%%%%%%%%%
%%%%%%%%%%%%%%%%%%%%%%%%%%%%%%%%%%%%%%%%%%%%%%%%%%%%%%%%%%%%%%%%%%%%
\subsection{Datasets}

We evaluate \ourmethod on the LineMod-Occluded (LMO)~\cite{lmo} and the YCB-Video (YCBV)~\cite{ycbv} datasets.

\noindent \textbf{LMO}~\cite{lmo}
contains RGBD images of real scenes with different configurations of objects placed on a table.
It provides the ground-truth 6D pose of eight of these objects, which are always present in the scene. 
Objects are poorly textured, of varying dimensions and placed in a cluttered scene, featuring a variety of lightning conditions.
We use the original test set of 1,213 real images, while for the training set the works we use as comparison use different combinations of synthetic and real images: the methods they use to generate the synthetic images and the number of samples for each type are not always clearly defined~\cite{pvn3d, ffb6d, e2ek, geometricaware6d}.
Differently, we only use the Photo Realistic Rendering (PBR) set of 50,000 synthetic images provided by the BOP challenge~\cite{bop-challenge} as it contains a large variety of pose configurations.
Following \cite{pvn3d}, we adopt an hole filling algorithm~\cite{hole_filling} to improve the depth quality on both training and test images.

\noindent \textbf{YCBV}~\cite{ycbv}
contains RGBD images of real scenes with different configurations of 21 objects taken from the YCB dataset~\cite{calli2015ycb}.
Objects have similar geometry (e.g.~boxes and cans) and are placed in various poses (e.g.~some objects are placed on top of others).
Unlike LMO, the objects are placed in different contexts.
We use the original test set of 20,738 real images.
As for LMO, state-of-the-art methods use different combinations of synthetic and real data~\cite{pvn3d, ffb6d, e2ek, geometricaware6d}.
For training, we choose 4,000 synthetic, 4,000 real, and 4,000 PBR images provided by the BOP challenge~\cite{bop-challenge} because we found that using only the PBR images leads to unsatisfactory results.
Also for YCBV we adopt a hole filling algorithm~\cite{hole_filling} on both train and test depth images as done in~\cite{pvn3d}.

\renewcommand{\arraystretch}{0.9}
\begin{table*}[t!]
\centering
\tabcolsep 4pt
\caption{
Comparison of RGB and RGBD methods performance on LMO~\cite{lmo} evaluated in terms of ADD(S)-0.1d.
Key:
$^\ast$: symmetric object,
DNNs: number of Deep Neural Networks used,
n.a.: information not available,
Det: object detections are used as prior,
Seg: object segmentation masks are used as prior,
\textbf{bold}: best result,
\underline{underline}: second best result.
}
\vspace{-3mm}
\label{tab:lmo_adds}
\resizebox{.8\textwidth}{!}{%
    \begin{tabular}{clccccccccccc}
        \toprule
        Input & Method & DNNs & Prior & Ape & Can & Cat & Drill & Duck & Eggbox$^\ast$ & Glue$^\ast$ & Holepuncher & Avg \\
        \midrule
        \multirow{2}{*}{\rotatebox{90}{RGB}} & SO-Pose \cite{sopose} & 1 & Det & 
            48.4 & 85.8 & 32.7 & 77.4 & 48.9 & 52.4 & 78.3 & 75.3 & 62.3 \\
        & ZebraPose \cite{zebrapose} & 8 & Det & 
            57.9 & 95.0 & 60.6 & 94.8 & 64.5 & 70.9 & 88.7 & 83.0 & 76.9 \\
        \midrule
        \multirow{8}{*}{\rotatebox{90}{RGBD}} & PVN3D \cite{pvn3d} & 1 & -- &
            33.9 & 88.6 & 39.1 & 78.4 & 41.9 & 80.9 & 68.1 & 74.7 & 63.2 \\
        & PR-GCN \cite{prgcn} & n.a. & Det &
            40.2 & 76.2 & 57.0 & 82.3 & 30.0 & 68.2 & 67.0 & 97.2 & 65.0 \\
        & FFB6D \cite{ffb6d} & 1 & -- &
            47.2 & 85.2 & 45.7 & 81.4 & 53.9 & 70.2 & 60.1 & 85.9 & 66.2 \\
        & DCL-Net \cite{dclnet} & n.a. & Det &
            56.7 & 80.2 & 48.1 & 81.4 & 44.6 & 83.6 & 79.1 & 91.3 & 70.6 \\
        & E2EK \cite{e2ek} & 8 & Seg &
            61.0 & 95.4 & 50.8 & 94.5 & 59.6 & 55.7 & 78.3 & 91.4 & 73.3 \\
        & Wu et al. \cite{geometricaware6d} & 8 & Det &
            66.1 & 97.4 & 70.7 & 95.4 & 70.1 & 61.2 & 59.8 & 95.7 & 77.1 \\
        & \ourmethod (ours) & 1 & -- & 
            65.4 & 96.7 & 64.8 & 97.8 & 71.7 & 54.1 & 83.2 & 97.9 & \underline{79.0} \\  % w/o heads, w/ TEASER
        & \ourmethod (ours) & 1 & Det & 
            63.6 & 94.8 & 63.4 & 97.4 & 73.4 & 74.6 & 80.4 & 97.3 & \textbf{80.6} \\  % w/o heads, w/ TEASER, Det.$^\dagger$
        \bottomrule
    \end{tabular}
}
\end{table*}
\renewcommand{\arraystretch}{1}

%%%%%%%%%%%%%%%%%%%%%%%%%%%%%%%%%%%%%%%%%%%%%%%%%%%%%%%%%%%%%%%%%%%%
%%%%%%%%%%%%%%%%%%%%%%%%%%%%%%%%%%%%%%%%%%%%%%%%%%%%%%%%%%%%%%%%%%%%
\subsection{Implementation details}

\noindent \textbf{LMO setting.}
Experiments on LMO share the following hyperparameters.
The input pair $(O, S)$ is first sampled to $V_O=$ 4,000 and $V_S=$ 50,000 points, respectively, and then quantised with a step of $Q=2$mm.
As feature extractor we use a MinkUNet34 \cite{minkowski} with output dimension $F=32$.
The correspondence estimation threshold used for the positive mining is $\tau_P=4$mm, and the maximum number of correspondences extracted is set to 1,000.
The safety threshold $\tau_{NO}$ is defined proportionally to the object $O$ diameter by setting $t_{\text{scale}}=0.1$ (see Fig.~\ref{fig:loss_detail}).
The hardest negative mining on $\mathcal{X}_O$ is performed in $\widetilde{\mathcal{N}}_i$.
When mining the hardest negatives on $\mathcal{X}_S$, instead of considering the full candidates set $\widetilde{\mathcal{N}}_j$ we randomly sample 10,000 points from it to reduce the spatial complexity.
HC loss margins are set as $\mu_P=0.1$, $\mu_N=10$, and coefficients are set to $\lambda_P = 1$, $\lambda_{NO} = 0.6$, and $\lambda_{NS} = 0.4$.
The feature extractor is trained on 50,000 PBR images for 12 epochs.
The pose is obtained by using the TEASER++~\cite{yang2020teaser} algorithm.

\noindent \textbf{YCBV setting.}
Experiments on YCBV share the same LMO hyperparameters except in the following cases.
We set $V_S=$ 20,000, as we found that it works on par with the original $V_S$ of LMO.
We believe this happens because YCBV objects are less occluded and their geometries are less complex than LMO objects.
As feature extractor we use a MinkUNet50 model \cite{minkowski}, trained on 12,000 mixed images for 110 epochs.
The pose is obtained with a RANSAC-based algorithm from Open3D~\cite{open3d}.
Experimentally, on YCBV we found that RANSAC yields better results than TEASER++.
We believe that this happens because TEASER++ is heavily based on correspondences~\cite{yang2020teaser} and for YCBV we use a lower resolution for the scene compared to LMO, which in turn reduces the number of correspondences.

%%%%%%%%%%%%%%%%%%%%%%%%%%%%%%%%%%%%%%%%%%%%%%%%%%%%%%%%%%%%%%%%%%%%
%%%%%%%%%%%%%%%%%%%%%%%%%%%%%%%%%%%%%%%%%%%%%%%%%%%%%%%%%%%%%%%%%%%%
\subsection{Evaluation metrics}

We use the ADD and ADD-S metrics that are defined as
% ==============================
\vspace{-2mm}
\begin{equation*}
\begin{aligned}
\text{ADD} &=
\frac{1}{V_O} \sum_{\textbf{x} \in \mathcal{X}_O} \left\lVert
(\textbf{R} \textbf{x} + \textbf{t}) - 
( \hat{\textbf{R}} \textbf{x} + \hat{\textbf{t}} ) 
\right\rVert,\\
\text{ADD-S} &=
\frac{1}{V_O} \sum_{\textbf{x}_1 \in \mathcal{X}_O} 
\min_{\textbf{x}_2 \in \mathcal{X}_O}
\left\lVert
(\textbf{R}\textbf{x}_1 + \textbf{t}) -
(\hat{\textbf{R}} \textbf{x}_2 + \hat{\textbf{t}})
\right\rVert,
\end{aligned}
\vspace{-2mm}
\end{equation*}
% ==============================
where $\mathbf{R}, \mathbf{t}$ and $\hat{\mathbf{R}}, \hat{\mathbf{t}}$ are the translation and rotation components of the predicted and the ground-truth poses of $\mathcal{X}_O \in \mathbb{R}^{V_O \times 3}$, respectively.
ADD(S) computes the ADD for non-symmetric objects and the ADD-S for symmetric ones.
Performance on LMO is assessed in term of the ADD(S)-0.1d metric~\cite{pvn3d, ffb6d, e2ek, geometricaware6d}, which computes the percentage of ADD(S) errors lower than $10\%$ of the object diameter~\cite{hodavn2016evaluation}.
Performance on YCBV is assessed in term of the ADD-S AUC metric~\cite{ycbv, pvn3d, ffb6d}.
The area-under-the-curve (AUC) of ADD-S is obtained by computing the cumulative percentage of ADD-S errors lower than a threshold varying from 1mm to 100mm.
Note that in ADD(S)-0.1d the success thresholds are relative to the object diameters, while in ADD-S AUC they are absolute.

%%%%%%%%%%%%%%%%%%%%%%%%%%%%%%%%%%%%%%%%%%%%%%%%%%%%%%%%%%%%%%%%%%%%
%%%%%%%%%%%%%%%%%%%%%%%%%%%%%%%%%%%%%%%%%%%%%%%%%%%%%%%%%%%%%%%%%%%%
\subsection{Quantitative results}

Tab.~\ref{tab:lmo_adds} reports the results on LMO~\cite{lmo} in term of ADD(S)-0.1d:
for completeness we added the two best performing RGB methods (top), while the other ones are RGBD methods (bottom).
As reported in the Prior column, most methods rely on additional priors, either in the form of object detections (Det) or of object segmentation masks (Seg).
\ourmethod outperforms all the other methods by a large margin without using any prior (penultimate row):
it outperforms Wu et al.~by 1.9\%, E2EK by 5.7\%, DCL-Net by 8.4\%, FFB6D by 12.8\%, PR-GCN by 14.0\%, and PVN3D by 15.8\%.
Note that Wu et al.~\cite{geometricaware6d} and E2EK~\cite{e2ek} train a different deep neural network for each object (DNNs column), whereas we train only a single deep neural network, saving learning parameters and training time.
Moreover, when we use the object detections obtained with YOLOv8~\cite{yolov8} (last row), the performance of \ourmethod further improves, outperforming Wu et al.~by 3.5\%, E2EK by 7.3\%, and all the other methods by more than 10.0\%.
Note that detectors are prone to errors: when detections are wrong, the object pose will be wrong too. 
We can observe that the detector is more effective with Duck and Eggbox.
The first is a particularly small object, therefore more likely to be occluded. 
The second undergoes frequent occlusions (other objects are on top of it in several images), thus making localisation difficult without a detector.
To further understand the negative impact of the detector, we compute the percentage of poses which are wrong when we use detections and correct when we do not use detections. 
For Ape, Can and Glue, this percentage is 3.3\%, 1.7\%, and 5.1\%, respectively. 
Please refer to the Supplementary Material for a comprehensive analysis of the detector impact.

\renewcommand{\arraystretch}{0.9}
\begin{table}[t]
\centering
\tabcolsep 3pt
\caption{
Performance of RGBD methods on YCBV~\cite{ycbv} evaluated in ADD-S AUC.
Key:
$^\ast$: symmetric object,
DNNs: number of Deep Neural Networks used,
Det: object detections are used as prior,
Seg: object segmentation masks are used as prior,
\textbf{bold}: best result,
\underline{underline}: second best result.
}
\vspace{-3mm}
\label{tab:ycbv_rgbd_addsauc}
\resizebox{\columnwidth}{!}{%
    \begin{tabular}{lccc|ccc}
        \toprule
        \multirow{2}{*}{Method} & PVN3D & FFB6D & \ourmethod & E2EK & Wu et al. & \ourmethod \\
        & \cite{pvn3d} & \cite{ffb6d} & (ours) & \cite{e2ek} & \cite{geometricaware6d} & (ours) \\
        \midrule
        DNNs & 1 & 1 & 1 & 21 & 21 & 1 \\
        Prior & -- & -- & -- & Seg & Det & Det \\
        \midrule
        master chef can &
            80.5 & 80.6 & 96.1 & 79.6 & 100.0 & 96.3 \\
        cracker box &
            94.8 & 94.6 & 96.4 & 95.1 &  98.8 & 96.7 \\
        sugar box &
            96.3 & 96.6 & 98.1 & 96.7 & 100.0 & 98.1 \\
        tomato soup can &
            88.5 & 89.6 & 93.1 & 89.8 &  97.5 & 95.8 \\
        mustard bottle &
            96.2 & 97.0 & 98.3 & 96.5 & 100.0 & 98.3 \\
        tuna fish can &
            89.3 & 88.9 & 82.8 & 90.7 &  99.9 & 97.6 \\
        pudding box &
            95.7 & 94.6 & 95.2 & 96.9 & 100.0 & 97.3 \\
        gelatin box &
            96.1 & 96.9 & 98.7 & 97.5 & 100.0 & 98.7 \\
        potted meat can &
            88.6 & 88.1 & 79.9 & 90.8 &  84.1 & 89.8 \\
        banana &
            93.7 & 94.9 & 98.3 & 94.4 & 100.0 & 98.3 \\
        pitcher base &
            96.5 & 96.9 & 97.9 & 95.6 & 100.0 & 97.9 \\
        bleach cleanser &
            93.2 & 94.8 & 95.9 & 94.0 &  99.9 & 96.7 \\
        bowl$^\ast$ &
            90.2 & 96.3 & 97.3 & 96.0 &  94.5 & 98.2 \\
        mug &
            95.4 & 94.2 & 97.4 & 95.3 & 100.0 & 97.7 \\
        power drill &
            95.1 & 95.9 & 98.2 & 96.6 & 100.0 & 98.2 \\ 
        wood block$^\ast$ &
            90.4 & 92.6 & 95.2 & 93.8 &  98.0 & 96.4 \\
        scissors &
            92.7 & 95.7 & 93.9 & 97.9 & 100.0 & 95.9 \\
        large marker &
            91.8 & 89.1 & 97.5 & 95.0 &  99.9 & 98.3 \\
        large clamp$^\ast$ &
            93.6 & 96.8 & 80.6 & 97.2 &  91.1 & 93.8 \\
        extra large clamp$^\ast$ &
            88.4 & 96.0 & 77.4 & 96.7 &  81.0 & 94.7 \\
        foam brick$^\ast$ &
            96.8 & 97.3 & 94.6 & 97.2 &  99.8 & 97.6 \\
        \midrule
        Avg & 91.8 & \underline{92.7} & \textbf{93.5} & 94.4 & \textbf{97.4} & \underline{96.8} \\ 
        \bottomrule
    \end{tabular}
}
\end{table}
\renewcommand{\arraystretch}{1}

Tab.~\ref{tab:ycbv_rgbd_addsauc} reports the results on YCBV~\cite{ycbv} in ADD-S AUC compared with other RGBD-based methods.
The row Prior indicates eventual additional priors used by each method.
The default configuration of \ourmethod does not require any input prior and uses a deep neural network for all the objects.
\ourmethod outperforms recent competitors that do not use input priors: it outperforms FFB6D by 0.8\% and PVN3D by 1.7\%.
E2EK~\cite{e2ek} and Wu et al.~\cite{geometricaware6d} instead consider input priors in the form of object segmentation masks and object detections, respectively, and train a model for each object (DNNs row).
When we use input priors in the form of detections, \ourmethod outperforms E2EK by $2.4\%$ and slightly underperforms Wu et al.~by $-0.6\%$.
We also observe that, thanks to multi-scale representation provided by the UNet, we obtain good performance also on symmmetric objects without the need of specific techniques to handle symmetry.
Note that we employed detections in both Tabs.~\ref{tab:lmo_adds}\&\ref{tab:ycbv_rgbd_addsauc} to illustrate their potential use in improving registration efficacy, though not obligatory.
Specifically in Tab.~2, when we compare with methods based on the same assumptions as ours, \ourmethod achieves state-of-the-art performance, see comparison with PVN3D \cite{pvn3d} and FFB6D \cite{ffb6d}.
When we compare with methods that use 21 models instead of 1 (as ours), we fall slightly behind the best (see comparison with E2EK \cite{e2ek} and Wu et al.~\cite{geometricaware6d}).

% ==================================================================
\begin{figure}[t]
    \centering
    \input{main/figures/qualitative/lmo_qualitative}
    \vspace{-5mm}
    \caption{
    Qualitative results on LMO~\cite{lmo}.
    Colour key:
    {\color{ape}{\Large$\bullet$}}~Ape,
    {\color{can}{\Large$\bullet$}}~Can,
    {\color{cat}{\Large$\bullet$}}~Cat,
    {\color{driller}{\Large$\bullet$}}~Drill,
    {\color{duck}{\Large$\bullet$}}~Duck,
    {\color{eggbox}{\Large$\bullet$}}~Eggbox,
    {\color{glue}{\Large$\bullet$}}~Glue,
    {\color{holepuncher}{\Large$\bullet$}}~Holepuncher.
    }
    \label{fig:lmo_qualitative}
\end{figure}
% ==================================================================

%%%%%%%%%%%%%%%%%%%%%%%%%%%%%%%%%%%%%%%%%%%%%%%%%%%%%%%%%%%%%%%%%%%%
%%%%%%%%%%%%%%%%%%%%%%%%%%%%%%%%%%%%%%%%%%%%%%%%%%%%%%%%%%%%%%%%%%%%
\subsection{Qualitative results}

Fig.~\ref{fig:lmo_qualitative} shows some examples of successes and failures on the test set of LMO dataset.
The upper row shows the ground-truth poses, and the bottom one shows the poses predicted by our model.
Note how FCGF6D is capable of estimating the correct pose even in case of partial objects (i.e. the glue in the first image).
However, our model fails in case of partial objects with ambiguities (the duck in the second image), or of atypical occlusions (the eggbox in the second image: the training set do not contain this degree of occlusions).

\begin{figure}[t]
    \centering
    \input{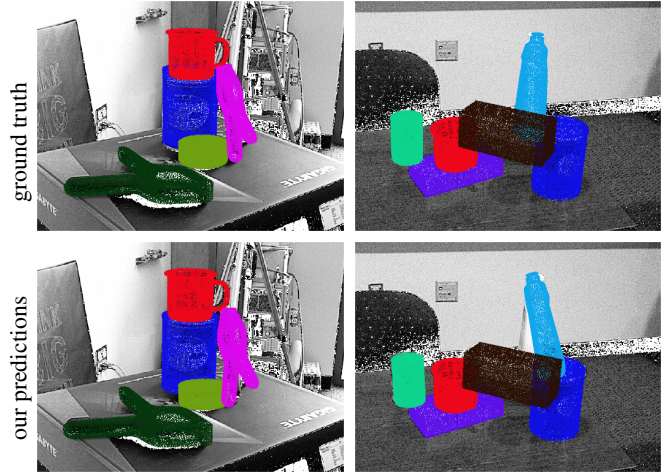}
    \vspace{-5mm}
    \caption{
    Qualitative results on YCBV~\cite{ycbv}.
    Colour key:
    {\color{master chef can}{\Large$\bullet$}}~master chef can,
    {\color{sugar box}{\Large$\bullet$}}~sugar box,
    {\color{tomato soup can}{\Large$\bullet$}}~tomato soup can,
    {\color{tuna fish can}{\Large$\bullet$}}~tuna fish can,
    {\color{bleach cleanser}{\Large$\bullet$}}~bleach cleanser,
    {\color{mug}{\Large$\bullet$}}~mug,
    {\color{wood block}{\Large$\bullet$}}~wood block,
    {\color{large clamp}{\Large$\bullet$}}~large clamp,
    {\color{extra large clamp}{\Large$\bullet$}}~extra large clamp.
    }
    \label{fig:ycbv_qualitative}
\end{figure}

Fig.~\ref{fig:ycbv_qualitative} shows some examples of successes and failures on the test set of YCBV. FCGF6D appears prone to rotation errors (the large clamp in the first image), especially in case of partially occluded objects (the bleach cleanser in the second image). However, the poses are generally accurate.

% ==================================================================
% ==================================================================

%%%%%%%%%%%%%%%%%%%%%%%%%%%%%%%%%%%%%%%%%%%%%%%%%%%%%%%%%%%%%%%%%%%%
%%%%%%%%%%%%%%%%%%%%%%%%%%%%%%%%%%%%%%%%%%%%%%%%%%%%%%%%%%%%%%%%%%%%
\subsection{Ablation study}

\renewcommand{\arraystretch}{0.9}
\begin{table}[t!]
\centering
\tabcolsep 3pt
\caption{
Ablation study on the Drill object of LMO.
Performance is compared in RRE [radians] and RTE [cm] errors, FMR and ADD(S)-0.1d (shortened to ADD) scores.
$\Delta$ shows the improvement of each contribution in terms of ADD(S)-0.1d with respect to the previous row.
}
\label{tab:ablation}
\vspace{-3mm}
\resizebox{\columnwidth}{!}{%
    \begin{tabular}{clrrrrr}
        \toprule
        & Improvements &
        RRE{\color{black!50}{$\,\downarrow$}} &
        RTE{\color{black!50}{$\,\downarrow$}} &
        FMR{\color{black!50}{$\,\uparrow$}} &
        ADD{\color{black!50}{$\,\uparrow$}} &
        $\Delta$ \\
        \toprule
        & Baseline & 2.2 & 9.6 & 0 & 0.2 & -- \\
        \midrule
        % Loss
        \multirow{2}{*}{\rotatebox{90}{Loss}} & $+$ $\tau_{NS} = 0.1 D_S$ & 1.8 & 12.2 & 0 & 0.4 & +0.2 \\
        & $+$ $\tau_{NS} = 0.1 D_O$ & 1.1 & 5.3 & 0.2 & 18.2 & +17.8 \\
        \midrule
        % Architecture
        \multirow{2}{*}{\rotatebox{90}{Arch.}} & $+$ Independent weights & 1.2 & 3.7 & 0 & 29.1 & +10.9 \\
        & $+$ Add RGB information & 0.6 & 2.2 & 38.5 & 63.3 & +34.2 \\
        \midrule
        % Data augmentation
        \multirow{2}{*}{\rotatebox{90}{Aug.}} & $+$ Colour augmentation & 0.6 & 2.2 & 32.0 & 65.4 & +2.1 \\
        & $+$ Random erasing & 0.3 & 1.8 & 78.4 & 75.6 & +10.2 \\
        \midrule
        % Optimization
        \multirow{3}{*}{\rotatebox{90}{Optim.}} & $+$ SGD $\to$ Adam & 0.1 & 1.1 & 93.4 & 95.8 & +20.2 \\
        & $+$ Adam $\to$ AdamW & 0.1 & 0.9 & 93.9 & 96.4 & +0.6 \\
        & $+$ Exp $\to$ Cosine & 0.1 & 0.9 & 93.6 & 96.6 & +0.2 \\
        \bottomrule
    \end{tabular}
}
\end{table}
\renewcommand{\arraystretch}{1}

% ==================================================================
\begin{figure}[t!]
\input{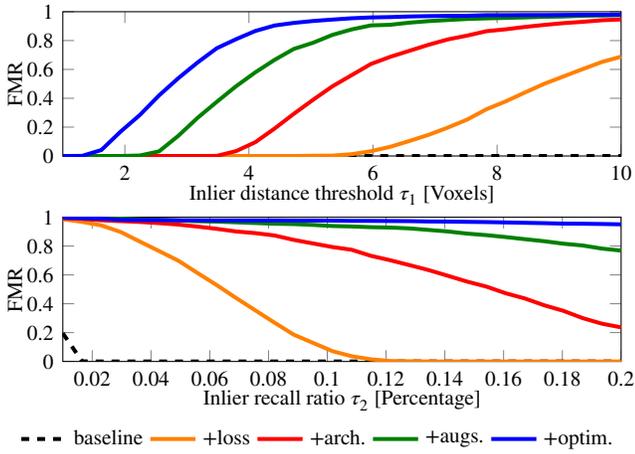}
\vspace{-8mm}
\caption{Feature Matching Recall (FMR) as a function of $\tau_1$ and $\tau_2$. 
When varying $\tau_1$ (top) we set $\tau_2$=5\%, and when varying $\tau_2$ (bottom) we set $\tau_1$=10 voxels.
}
\label{fig:fmr}
\end{figure}
% ==================================================================

We conduct an ablation study on the Drill object of the LMO dataset by training FCGF6D for five epochs.
We choose the closest setting to FCGF as baseline: 
no safety threshold in the loss, 
shared network weights, 
no RGB information, 
SGD optimiser with $\text{lr}_{\text{init}}=10^{-3}$,
exponential scheduler with $\gamma = 0.99$.
We perform a single experiment for each added component to assess their individual contribution.
As metrics, we use ADD(S), Relative Rotation Error (RRE), Relative Translational Error (RTE), and Feature Matching Recall (FMR) \cite{FCGF,GEDI}.
RRE and RTE show how the two pose components (rotation and translation) are affected.
FMR indirectly measures the number of iterations required by a registration algorithm, e.g.~RANSAC, to estimate the transformation between two point clouds.
We set the inlier distance threshold as $\tau_1$ = 5 voxels, and the inlier recall ratio as $\tau_2$ = 5\%.

Tab.~\ref{tab:ablation} shows that the largest contributions in ADD(S)-0.1d are: introducing the safety threshold in the loss (+17.8), adding RGB information (+34.2), and adopting Adam optimiser (+20.2). 
We also note that the gain in ADD(S)-0.1d is not always consistent with the FMR: when RGB augmentation is added, there is a gain in ADD(S)-0.1d of 2.1, but the FMR drops by 6.5.
A more detailed analysis of FMR with different values of $\tau_1$ and $\tau_2$ is shown in Fig.~\ref{fig:fmr}.

\subsection{Training and inference time}

The training time is about one week for each dataset using two NVIDIA A40 GPUs.
Tab.~\ref{tab:times} reports the comparison of the number of parameters, inference GPU memory footprint, and inference time (using a GeForce RTX 3050 GPU) on YCBV. 
We were unable to test E2EK \cite{e2ek} as the code is unavailable, whereas we used the authors' original code for the other papers.
\ourmethod has a significantly smaller memory footprint than the main competitors, and the inference time is comparable. 
In a scenario where multiple objects are expected, our closest competitor~\cite{geometricaware6d} uses a different model for each object, thereby requiring more memory. 
Our method requires less memory because we train only a single model.
Note that using the whole scene as input is advantageous in a practical scenario where $N$ instances of the same object are present. 
Here, we need a single forward pass, followed by $N$ registrations. 
Instead, methods that rely on image crops~\cite{prgcn,dclnet, geometricaware6d} require a forward pass for each instance.

\tabcolsep 4pt
\renewcommand{\arraystretch}{0.7}
\begin{table}[]
\centering
\caption{Inference time and memory footprint. 
Time is for a single image, and includes network inference (inf.) and registration (reg.) times. $N$ is the number of trained models.}
\vspace{-3mm}
\resizebox{\columnwidth}{!}{
    \begin{tabular}{lcccl}
        \toprule
        Method & DNNs & Params [M] & Memory [GB] & Time [ms] (inf.+reg.) \\
        \toprule
        PVN3D \cite{pvn3d} & 1 & 38.6 & 3.17 & 417 (154 + 263) \\
        FFB6D \cite{ffb6d} & 1 & 33.8 & 2.46 & 285 (146 + 139) \\
        Wu et al.~\cite{geometricaware6d} & $N$ & 23.8$\times N$ & 2.04$\times N$ & 144 (143 + 1) \\
        \midrule
        Ours & 1 & 63.5 & 1.3 & 156 (118 + 38) \\
        \bottomrule
    \end{tabular}
\label{tab:times}
}
\end{table}

%%%%%%%%%%%%%%%%%%%%%%%%%%%%%%%%%%%%%%%%%%%%%%%%%%%%%%%%%%%%%%%
%%%%%%%%%%%%%%%%%%%%%%%%%%%%%%%%%%%%%%%%%%%%%%%%%%%%%%%%%%%%%%%
%%%%%%%%%%%%%%%%%%%%%%%%%%%%%%%%%%%%%%%%%%%%%%%%%%%%%%%%%%%%%%%
\section{Conclusions}\label{sec:conclusions}

We revisited the Fully Convolutional Geometric Feature (FCGF) approach to tackle the problem of object 6D pose estimation.
FCGF uses sparse convolutions to learn point-wise features while optimising a hardest contrastive loss. 
Key modifications to the loss, input data representations, training strategies, and data augmentations to FCGF enabled us to outperform competitors on popular benchmarks. 
A thorough analysis is conducted to study the contribution of each modification to achieve state-of-the-art performance.
Future research directions include the application of our approach to generalisable 6D pose estimation~\cite{onepose}.

\noindent\textbf{Limitations.}
Minkowski engine is computational efficient but has a large memory footprint at training time.
We mitigated this by downsampling the scene point cloud and by adopting quantisation. It would be interesting to understand how not to lose the input point cloud resolution while maintaining a modest memory footprint.

\small{
\section*{Acknowledgements}
\label{sec:acknowledgements}
We are grateful to Andrea Caraffa for his support with the computation of the detection priors and to Nicola Saljoughi for his contributions during the early stage of the project.

This work was supported by the European Union’s Horizon Europe research and innovation programme under grant agreement No 101058589 (AI-PRISM), and by the PNRR project FAIR - Future AI Research (PE00000013), under the NRRP MUR program funded by the NextGenerationEU.
}

{\small
\bibliographystyle{main/ieee_fullname}
\bibliography{main}
}

\end{document}

\title{Supplementary material for\\ Revisiting Fully Convolutional Geometric Features for Object 6D Pose Estimation}

\author{
\begin{minipage}{0.32\textwidth}
    \vspace*{-7mm}
    \centering
    Jaime Corsetti
\end{minipage}
\begin{minipage}{0.32\textwidth}
    \vspace*{-7mm}
    \centering
    Davide Boscaini
\end{minipage}
\begin{minipage}{0.32\textwidth}
    \vspace*{-7mm}
    \centering
    Fabio Poiesi
\end{minipage} \\
\begin{minipage}{0.32\textwidth}
    \vspace*{-3mm}
    \centering
    {\tt\small jaime.corsetti98@gmail.com}
\end{minipage}
\begin{minipage}{0.32\textwidth}
    \vspace*{-3mm}
    \centering
    {\tt\small dboscaini@fbk.eu}
\end{minipage}
\begin{minipage}{0.32\textwidth}
    \vspace*{-3mm}
    \centering
    {\tt\small poiesi@fbk.eu}
\end{minipage} \\
\begin{minipage}{1.0\textwidth}
    \vspace*{1mm}
    \centering
    Fondazione Bruno Kessler, Italy
\end{minipage}
}

\maketitle

\section{Introduction}
\label{sec:intro}

We provide additional material in support of our main paper.
The material is organised as follows:

\setlist{nolistsep}
\begin{itemize}
    \item In Sec.~\ref{sec:qualitative} we show qualitative results on the LMO~\cite{lmo} and YCBV~\cite{ycbv} datasets, divided into success (Figs.~\ref{fig:lmo_success} \& \ref{fig:ycbv_success}) and failure (Figs.~\ref{fig:lmo_fail} \& \ref{fig:ycbv_fail}) cases. 
    We also report examples of how pose registration is affected by correct and incorrect detections.
    To assess the quality of the learnt features, we visualise the distances in the feature space between a point and all the other points (Figs.~\ref{fig:feat_all_lmo} \& \ref{fig:feat_all_ycbv}).
    \item In Sec.~\ref{sec:ablation_ycbv} we report an ablation study on one object of the YCBV dataset (Tab.~\ref{tab:ablation_ycbv}).
    \item In Sec.~\ref{sec:add_ablation_lmo} we report an additional ablation study on the choice of the $t_\text{scale}$ hyperparameter.
    \item In Sec.~\ref{sec:detections} we show the detection metrics on the test set used in our results (Tabs.~\ref{tab:detection_lmo} \& \ref{tab:detection_ycbv}) and highlight problems we found in the ground-truth annotations (Fig.~\ref{fig:scissors}). 
    We also report the percentage of cases in which the detector causes a failure or a success in the pose registration.
\end{itemize}

\section{Additional qualitative results}
\label{sec:qualitative}

In Figs.~\ref{fig:lmo_success} \& \ref{fig:ycbv_success} we show examples of correct pose predictions on LMO~\cite{lmo} and YCBV~\cite{ycbv}, respectively.
The top row shows the ground-truth poses, while the bottom row shows the results obtained with our experimental setting.

In Figs.~\ref{fig:lmo_fail} \& \ref{fig:ycbv_fail} we show examples of wrong pose predictions on LMO~\cite{lmo} and YCBV~\cite{ycbv}, respectively. 
To highlight the detector contribution, we show the ground-truth poses (top row), our results with the object detection prior (middle row), and our results without the object detection prior (bottom row).

In LMO~\cite{lmo} we observe that the detector appears to be strongly influenced by the colour of the object, as it confuses the Can, Eggbox, and Glue objects which show similar colours (see Fig.~\ref{fig:lmo_fail}(a-b-c)).

Something similar occurs in Fig.~\ref{fig:lmo_fail}(d), where the pose prediction for Holepuncher is correct when the object detector prior is not used, and wrong when it is used.
This is due to the colour similarity between Holepuncher and the toy car behind the Glue object.

In YCBV~\cite{ycbv} the object detector handles important errors (Figs.~\ref{fig:ycbv_fail}(b-c)) and improves the pose prediction accuracy (Figs.~\ref{fig:ycbv_fail}(f-g-h)). 
We also show a failure case where a wrong detection causes an inaccurate pose prediction (the extra large clamp object in Fig.~\ref{fig:ycbv_fail}(e)).

\begin{figure*}[t]
    \centering
    \input{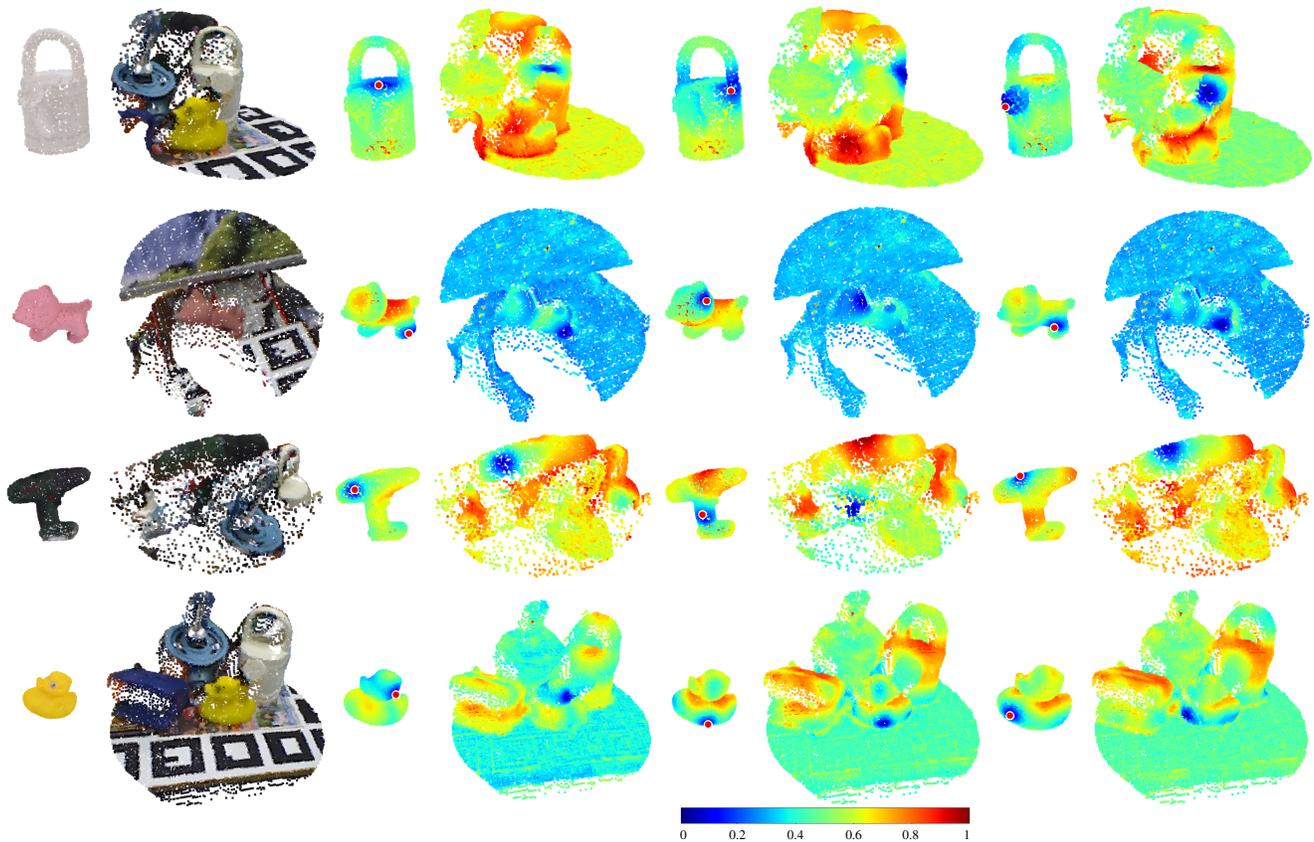}
    \smallskip
    \caption{
    Qualitative evaluation of the features learned with our approach on the Can, Duck, Drill, and Cat objects of the LMO dataset~\cite{lmo}. We show the Euclidean distance between the features of a reference point (red point) on the object $O$ and the features computed both on the rest of the points on $O$ and on the scene $S$ (cropped around the object of interest to facilitate visualisation).
    Cold and hot colours represent small and large distances, respectively.
    Ideal descriptors would produce a distance map with a sharp minimum at the corresponding point and no spurious local minima at other locations. 
    }
    \label{fig:feat_all_lmo}
\end{figure*}

\begin{figure*}[t]
    \centering
    \input{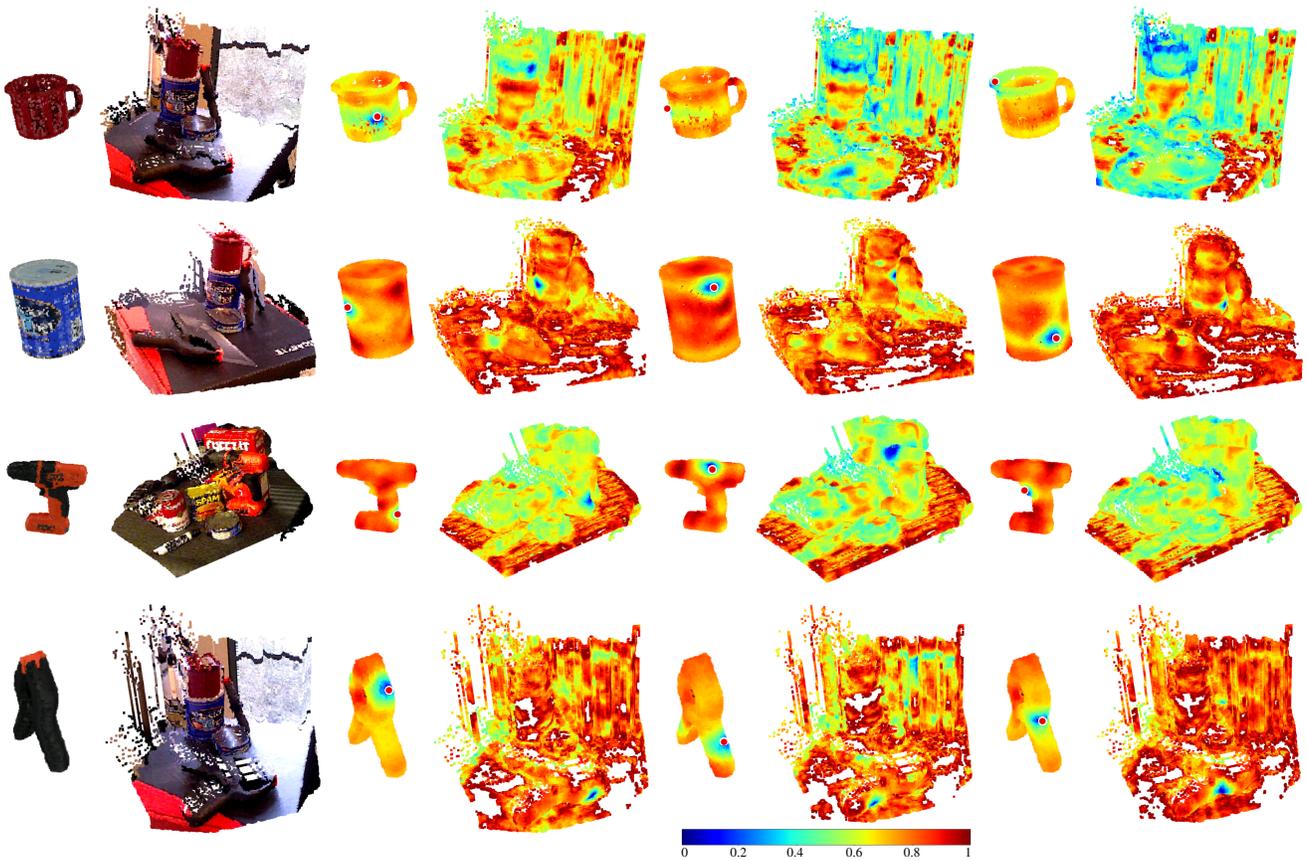}
    \smallskip
    \caption{
    Qualitative evaluation of the features learned with our approach on the mug, master chef can, power drill, and extra large clamp objects of the YCBV dataset~\cite{ycbv}. We show the Euclidean distance between the features of a reference point (red point) on the object $O$ and the features computed both on the rest of the points on $O$ and on the scene $S$ (cropped around the object of interest to facilitate visualisation).
    Cold and hot colours represent small and large distances, respectively.
    Ideal descriptors would produce a distance map with a sharp minimum at the corresponding point and no spurious local minima at other locations.
    }
    \label{fig:feat_all_ycbv}
\end{figure*}

To examine the point-level features learned by FCGF6D, we select pairs of point clouds and visualise the distance in the feature space of each point with respect to a reference point.
Consider Figs.~\ref{fig:feat_all_lmo} \& \ref{fig:feat_all_ycbv} for LMO and YCBV, respectively.
Given a pair $(O, S)$, respectively object and scene, we randomly select three points belonging to a correspondence on $O$, then we compute the distance in the feature space of each point in $O$ and $S$ from it.
The distance is then normalised for better visualisation.
We show the RGB point clouds on the left, and the input pairs with the feature distance on the right.
The reference point is depicted as a red point on $O$.

In LMO~\cite{lmo} (Fig.~\ref{fig:feat_all_lmo}), we can observe that the distance in the scene point clouds is small near the point corresponding to the reference one. 
We can also observe that the model can learn a certain degree of symmetry: in the second visualisation of the Can object, the distance in feature space of a point which is symmetric to the reference one is small.

In Fig.~\ref{fig:feat_all_ycbv} we show the visualisation on YCBV~\cite{ycbv}.
In general the distribution of the distances appears to be noisier and less smooth than in LMO. 
We believe this to be due to the stronger sampling on $S$ that we perform on YCBV, unlike the one done on LMO (20,000 points against 50,000 points).
Another possible cause is that the RGB test images of YCBV are of a lower quality than the ones of LMO.
As in the LMO visualisation, the corresponding point of the reference point in the scene has a small distance in the feature space.
We can also observe the resulting distance in the case of similar objects: in the rightmost case of the extra large clamp, one of the points on a similar object (the large clamp) is similar to the reference one in the feature space.
Because our features are learned to describe a local patch, their quality can be limited by the presence of similar geometric structures in other objects.
Also in this case we can observe how symmetry influences the feature space, by considering the visualisation of the mug object.
We can observe that, especially in the second and third pairs, the most similar points on $S$ are the ones on the radial symmetry axis of $O$.

\begin{figure*}[t]
    \input{supp/figures/lmo_qualitative/success}
    \vspace*{-5mm}
    \caption{
    Example of success cases on the LMO dataset~\cite{lmo}. Colour key:
    {\color{ape}{\Large$\bullet$}}~Ape,
    {\color{can}{\Large$\bullet$}}~Can,
    {\color{cat}{\Large$\bullet$}}~Cat,
    {\color{driller}{\Large$\bullet$}}~Drill,
    {\color{duck}{\Large$\bullet$}}~Duck,
    {\color{eggbox}{\Large$\bullet$}}~Eggbox,
    {\color{glue}{\Large$\bullet$}}~Glue,
    {\color{holepuncher}{\Large$\bullet$}}~Holepuncher.
    }
    \label{fig:lmo_success}
\end{figure*}

\begin{figure*}[t]
    \input{supp/figures/lmo_qualitative/fail}
    \vspace*{-5mm}
    \caption{
    Example of failure cases on the LMO dataset~\cite{lmo}. Colour key:
    {\color{ape}{\Large$\bullet$}}~Ape,
    {\color{can}{\Large$\bullet$}}~Can,
    {\color{cat}{\Large$\bullet$}}~Cat,
    {\color{driller}{\Large$\bullet$}}~Drill,
    {\color{duck}{\Large$\bullet$}}~Duck,
    {\color{eggbox}{\Large$\bullet$}}~Eggbox,
    {\color{glue}{\Large$\bullet$}}~Glue,
    {\color{holepuncher}{\Large$\bullet$}}~Holepuncher.
    }
    \label{fig:lmo_fail}
\end{figure*}

\begin{figure*}[t]
    \input{supp/figures/ycbv_qualitative/success}
    \caption{
    Example of success cases on the YCBV dataset~\cite{ycbv}. Colour key:
    {\color{master chef can}{\Large$\bullet$}}~master chef can,
    {\color{cracker box}{\Large$\bullet$}}~cracker box,
    {\color{sugar box}{\Large$\bullet$}}~sugar box,
    {\color{tomato soup can}{\Large$\bullet$}}~tomato soup can,
    {\color{mustard bottle}{\Large$\bullet$}}~mustard bottle,
    {\color{tuna fish can}{\Large$\bullet$}}~tuna fish can,
    {\color{pudding box}{\Large$\bullet$}}~pudding box,
    {\color{gelatin box}{\Large$\bullet$}}~gelatin box,
    {\color{potted meat can}{\Large$\bullet$}}~potted meat can,
    {\color{banana}{\Large$\bullet$}}~banana,
    {\color{pitcher base}{\Large$\bullet$}}~pitcher base,
    {\color{bleach cleanser}{\Large$\bullet$}}~bleach cleanser,
    {\color{bowl}{\Large$\bullet$}}~bowl,
    {\color{mug}{\Large$\bullet$}}~mug,
    {\color{power drill}{\Large$\bullet$}}~power drill,
    {\color{wood block}{\Large$\bullet$}}~wood block,
    {\color{scissors}{\Large$\bullet$}}~scissors,
    {\color{large marker}{\Large$\bullet$}}~large marker,
    {\color{large clamp}{\Large$\bullet$}}~large clamp,
    {\color{extra large clamp}{\Large$\bullet$}}~extra large clamp,
    {\color{foam brick}{\Large$\bullet$}}~foam brick.
    }
    \label{fig:ycbv_success}
\end{figure*}

\begin{figure*}[t]
    \input{supp/figures/ycbv_qualitative/fail}
    \vspace*{-5mm}
    \caption{
    Example of failure cases on the YCBV dataset~\cite{ycbv}. Colour key:
    {\color{master chef can}{\Large$\bullet$}}~master chef can,
    {\color{cracker box}{\Large$\bullet$}}~cracker box,
    {\color{sugar box}{\Large$\bullet$}}~sugar box,
    {\color{tomato soup can}{\Large$\bullet$}}~tomato soup can,
    {\color{mustard bottle}{\Large$\bullet$}}~mustard bottle,
    {\color{tuna fish can}{\Large$\bullet$}}~tuna fish can,
    {\color{pudding box}{\Large$\bullet$}}~pudding box,
    {\color{gelatin box}{\Large$\bullet$}}~gelatin box,
    {\color{potted meat can}{\Large$\bullet$}}~potted meat can,
    {\color{banana}{\Large$\bullet$}}~banana,
    {\color{pitcher base}{\Large$\bullet$}}~pitcher base,
    {\color{bleach cleanser}{\Large$\bullet$}}~bleach cleanser,
    {\color{bowl}{\Large$\bullet$}}~bowl,
    {\color{mug}{\Large$\bullet$}}~mug,
    {\color{power drill}{\Large$\bullet$}}~power drill,
    {\color{wood block}{\Large$\bullet$}}~wood block,
    {\color{scissors}{\Large$\bullet$}}~scissors,
    {\color{large marker}{\Large$\bullet$}}~large marker,
    {\color{large clamp}{\Large$\bullet$}}~large clamp,
    {\color{extra large clamp}{\Large$\bullet$}}~extra large clamp,
    {\color{foam brick}{\Large$\bullet$}}~foam brick.
    }
    \label{fig:ycbv_fail}
\end{figure*}

\section{Ablation study on YCBV}
\label{sec:ablation_ycbv}

In Tab.~\ref{tab:ablation_ycbv} we report the results of our ablation study on YCBV~\cite{ycbv}.
We choose the large marker object and train a single model on it for each modification we applied.
Each model is trained for 20 epochs on the standard training set.
For the computation of the Feature Matching Recall (FMR), we set the distance threshold $\tau_1=10$ voxels and the inlier ratio threshold $\tau_2=5$\%, to account for the different density of the scene point cloud in YCBV.
All the other settings and parameters are the same as those in our ablation study on LMO~\cite{lmo} in the main paper.

We can observe that some changes do not increment performance, but instead cause a slight drop, in particular when adapting the safety threshold to the object dimension (third row, $-0.4$) and when colour augmentation is applied (sixth row, $-$0.3).
These additions do not benefit this particular object, but are instead advantageous when averaging all the object in the dataset.

We can note that, as in the ablation study on the LMO dataset in the main paper, the most significant improvements in ADD-S AUC result from applying the safety threshold ($+$1.5), adding RGB information ($+$5.5), and using the Adam optimiser ($+$12.3).
\renewcommand{\arraystretch}{0.9}
\begin{table}%[t!]
\centering
\tabcolsep 3pt
\caption{
Ablation study on the large marker object of YCBV.
Performance are compared in terms of RRE [radians] and RTE [cm] errors (the lower the better), and FMR and ADD-S AUC (shortened to ADD) scores (the higher the better).
$\Delta$ shows the improvement of each contribution in terms of ADD-S AUC with respect to the previous row.
}
\vspace{-3mm}
\resizebox{\columnwidth}{!}{%
\begin{tabular}{clrrrrr}
\toprule
& Improvements &
RRE{\color{black!50}{$\,\downarrow$}} &
RTE{\color{black!50}{$\,\downarrow$}} & 
FMR{\color{black!50}{$\,\uparrow$}} & 
ADD{\color{black!50}{$\,\uparrow$}} & 
$\Delta$ \\ 
\toprule
& Baseline & 2.0 & 4.6 & 0.00 & 77.2 & -- \\
\midrule
\multirow{2}{*}{\rotatebox{90}{Loss}} & $+$ $\tau_{NS} = 0.1 D_S$ & 2.0 & 4.2 & 0.00 & 78.7 & $+$1.5 \\
& $+$ $\tau_{NS} = 0.1 D_O$ & 2.0 & 4.3 & 0.00 & 78.3 & $-$0.4 \\
\midrule
\multirow{2}{*}{\rotatebox{90}{Arch.}} & $+$ Independent weights & 2.0 & 4.1 & 0.00 & 79.4 & $+$1.1 \\
& $+$ Add RGB information & 1.2 & 3.2 & 49.1 & 84.9 & $+$5.5 \\
\midrule
\multirow{2}{*}{\rotatebox{90}{Aug.}} & $+$ Color augmentation & 1.2 & 3.3 & 50.0 & 84.6 & $-$0.3 \\
& $+$ Random erasing & 1.2 & 3.1 & 53.4 & 85.2 & $+$0.6 \\
\midrule
\multirow{2}{*}{\rotatebox{90}{Optim.}} & $+$ SGD $\to$ Adam & 0.0 & 0.4 & 100 & 97.5 & $+$12.3 \\
& $+$ Adam $\to$ AdamW  & 0.0 & 0.4 & 100 & 97.5 & 0 \\
& $+$ Exp $\to$ Cosine & 0.0 & 0.4 & 100 & 97.4 & $-$0.1 \\\bottomrule
\end{tabular}}
\label{tab:ablation_ycbv}
\end{table}
\renewcommand{\arraystretch}{1}

\section{Additional ablation study on LMO}
\label{sec:add_ablation_lmo}
We include in Tab.~\ref{tab:tscale} an ablation study on the $t_\text{scale}$ hyperparameter, which is used to set the radius of the ball volume in which negative mining around a certain point is not allowed. We train on the Can object of LMO using the standard setting, and varying only $t_\text{scale}$.
We can observe that our choice of $t_\text{scale} = 0.1$ leads to the best result. When $t_\text{scale}$ is increased, many candidate points are forbidden to be used as negatives, therefore decreasing the final performance. On the other hand, a lower $t_\text{scale}$ implies negative pairs composed by points which are near in the 3D space. This reduces the performance, as similar points are forced to have different descriptors. Notably, the worst results is obtained when $t_\text{scale} = 0$, i.e. when no negative candidates are excluded.

\begin{table}
\tabcolsep 3pt
\caption{
Ablation study on the Can object of LMO. Performance is shown in terms of ADD-0.1 (the higher the better) in function of the hyperparameter $t_\text{scale}$.}
\centering
\resizebox{.9\columnwidth}{!}{
\begin{tabular}{c|ccccc}
    \toprule
    $t_\text{scale}$ & 0.0 & 0.01 & 0.05 & \textbf{0.1} & 0.5 \\
    ADD-0.1d & 66.55 & 91.80 & 93.79 & \textbf{93.95} & 81.28 \\
    \bottomrule
\end{tabular}
\label{tab:tscale}
}
\end{table}

\section{Contribution of the object detection prior}
\label{sec:detections}

\begin{table}%[t!]
\centering
\tabcolsep 4pt
\caption{
Object detector performance on LMO~\cite{lmo}. On this dataset, failures and successes of $\Delta_{S \to F}$ and $\Delta_{F \to S}$ are measured in terms of ADD(S)-0.1d. Key: $^\ast$: symmetric object.
}
\label{tab:detection_lmo}
\vspace{-3mm}
% \resizebox{1.0\columnwidth}{!}{
    \begin{tabular}{lcccc}
        \toprule
        Object & AP & $\text{AP}_{50}$ & $\Delta_{S \to F}$ & $\Delta_{F \to S}$ \\
        \midrule
        Ape & 64.9 & 95.8 & 3.3 & 1.5\\
        Can & 82.8 & 99.3 & 1.9 & 0 \\
        Cat & 69.6 & 90.8 & 1.5 & 0.1 \\
        Driller & 80.0 & 98.7 & 0.8 & 0.4 \\
        Duck & 77.9 & 98.2 & 0.7 & 2.4 \\
        Eggbox$^\ast$ & 58.8 & 85.9 & 4.3 & 24.8 \\
        Glue$^\ast$ & 55.1 & 87.3 & 5.1 & 2.3\\
        Holepuncher & 82.3 & 99.5 & 2.0 & 1.4\\
        \midrule
        Average & 71.4 & 94.4 & 1.2 & 2.7 \\
        \bottomrule
    \end{tabular}
% }
\end{table}

\renewcommand{\arraystretch}{0.9}
\begin{table}%[t!]
\centering
\tabcolsep 4pt
\caption{
Object detector performance on YCBV~\cite{ycbv}. On this dataset, failures and successes of $\Delta_{S \to F}$ and $\Delta_{F \to S}$ are measured in terms of ADD-S-0.1d. Key: $^\ast$: symmetric object.
}
\label{tab:detection_ycbv}
\vspace{-3mm}
% \resizebox{1.0\columnwidth}{!}{
    \begin{tabular}{lcccc}
        \toprule
        Object & AP & $\text{AP}_{50}$ & $\Delta_{S \to F}$ & $\Delta_{F \to S}$ \\
        \midrule
            master chef can & 89.8 & 99.5 & 0.0 & 0.5 \\
            cracker box & 91.5 & 99.5 & 0.1 & 0.3 \\
            sugar box & 96.0 & 99.5 & 0.0 & 0.0 \\
            tomato soup can & 85.3 & 98.8 & 0.0 & 5.1 \\
            mustard bottle & 97.0 & 99.5 & 0.0 & 0.0 \\
            tuna fish can & 88.5 & 99.5 & 1.9 & 18.3 \\
            pudding box & 96.7 & 99.5 & 3.0 & 8.8 \\
            gelatin box & 96.6 & 99.5 & 0.0 & 0.2 \\
            potted meat can & 84.2 & 99.5 & 0.7 & 7.8 \\
            banana & 90.9 & 99.5 & 0.0 & 0.1 \\
            pitcher base & 99.4 & 99.5 & 0.0 & 0.0 \\
            bleach cleanser & 88.3 & 99.4 & 0.2 & 1.6 \\
            bowl* & 93.9 & 99.5 & 1.7 & 3.4 \\
            mug & 91.5 & 99.5 & 0.0 & 0.4 \\
            power drill & 95.4 & 99.5 & 0.0 & 0.0 \\
            wood block* & 84.1 & 99.5 & 0.2 & 2.0 \\
            scissors & 21.3 & 27.6 & 0.2 & 2.3 \\
            large maker & 82.3 & 99.5 & 0.0 & 0.7 \\
            large clamp* & 80.0 & 95.4 & 0.7 & 10.6 \\
            extra large camp* & 83.0 & 97.8 & 1.5 & 20.1 \\
            foam brick* & 86.3 & 99.5 & 0.2 & 4.0 \\
        \midrule
            Average & 86.8 & 95.8 & 0.4 & 4.2 \\
        \bottomrule
    \end{tabular}
% }
\end{table}
\renewcommand{\arraystretch}{1}

We report in Tabs.~\ref{tab:detection_lmo} \& \ref{tab:detection_ycbv} the performance of the YOLOv8 detector~\cite{Yolov8} on the test set of LMO~\cite{lmo} and YCBV~\cite{ycbv}, respectively.
Following \cite{coco}, we report the area-under-the-curve of the Average Precision, obtained by varying the IoU threshold with respect to a ground truth detection from 0.5 to 0.95 with a step of 0.05 ($\text{AP}$).
We also report the recall on Average Precision, obtained with a fixed IoU threshold of 0.5 ($\text{AP}_{50}$).
We also measure how the performances in ADD(S)-0.1d and ADD-S-0.1d change when using a detector.

\begin{figure*}[!h]
\centering 
    \includegraphics[width=1.0\textwidth]{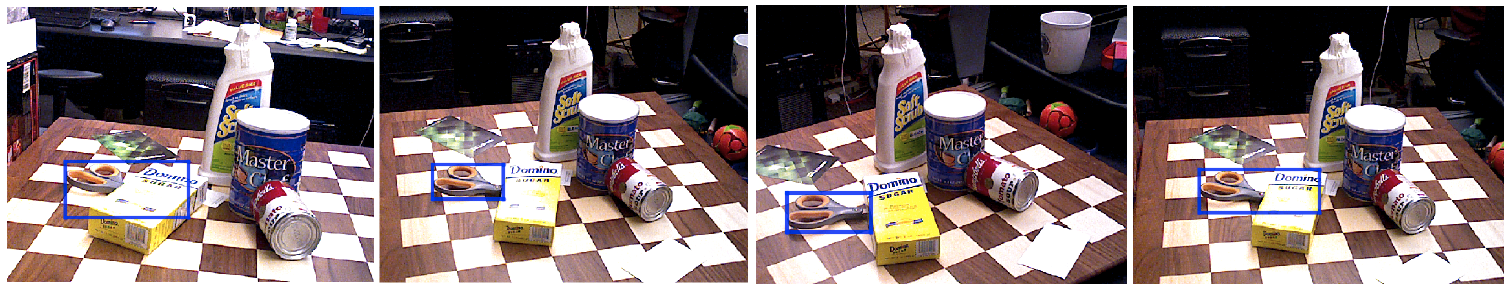}
    \caption{
    Examples of noisy ground truth detections of the scissors object of YCBV \cite{ycbv} provided by the BOP challenge~\cite{bop-challenge}.
    These annotations are used to train our YOLOv8~\cite{Yolov8} object detector.
    }
    \label{fig:scissors}
\end{figure*}

Because LMO and YCBV use ADD(S)-0.1d and ADD-S AUC, respectively, to measure the performance change, we consider ADD(S)-0.1d for LMO and in ADD-S-0.1d for YCBV.
As in the original definition~\cite{hodavn2016evaluation}, a success for the ADD-S-0.1d metric is obtained when the ADD-S error is below $0.1 D_O$, where $D_O$ is the object diameter, otherwise it is considered a failure.
The same applies to ADD(S)-0.1d.
Therefore, we define the following metrics:

\begin{itemize}
    \item  $\Delta_{S \to F}$ : the percentage of object instances for which there is a success when not using a detector and a failure when using it.
    \item $\Delta_{F \to S}$ : the percentage of object instances for which there is a failure when not using a detector and a success when using it.
\end{itemize}

In Tab.~\ref{tab:detection_lmo} we report the metrics on LMO~\cite{lmo}.
We can observe that the detector performance is correlated with the size of the object: Ape, Cat and Duck are very small and often occluded, and therefore perform worse than Can, Holepuncher, and Drill which are bigger.
The lower performance on Glue and Eggbox objects occurs because they are often confused by the detector. 
An example of this can be observed in  Fig.~\ref{fig:lmo_fail}(b).
This behaviour is also present with the Can object in Fig.~\ref{fig:lmo_fail}(a).
Despite these errors, we note that the Eggbox object greatly benefits from the detector (+24.8 in $\Delta_{F \to S}$), while for the other objects the performance gain is less significant.

In Tab.~\ref{tab:detection_ycbv} we show the detection performances on YCBV~\cite{ycbv}.
Unlike LMO, the introduction of the object detector is beneficial for all the objects, as the $\Delta_{F \to S}$ is always higher then the $\Delta_{S \to F}$ (i.e.~the detector solves more pose errors then it introduces for every object).
We can note how the detector helps in solving the problem of object similarity: the large clamp and extra large clamp objects are amongst the ones that benefit the most from it. 
As an example, in Fig.~\ref{fig:ycbv_fail}(a) we show how the registration of the two objects (large clamp in dark blue, extra large clamp in dark green) differs depending on the use of the detector. 
With prior detection (second row) the two clamps are registered correctly.
Without the detection (third row) the model registers both of them on the pose of the extra large clamp.
We observe in Tab.~\ref{tab:detection_ycbv} that the scissors object appears to be an outlier in the detector performance (AP$_{50}$ of 27.6 against an average of 95.8).
By examining the ground-truth detections we observed that they are noisy in the case of occlusion: sometimes the image portion where the object should be is also included in the bounding box, even if the object is not visible. 
See Fig.~\ref{fig:scissors} for an example. 
Despite this, the scissors performance in pose estimation still benefits from the detector.

{\small
\bibliographystyle{supp/ieee_fullname}
\bibliography{supp}
}